\pdfoutput=1
\documentclass[pdflatex,sn-mathphys-num]{sn-jnl}
\usepackage{graphicx}%
\usepackage{multirow}%
\usepackage{amsmath,amssymb,amsfonts}%
\usepackage{amsthm}%
\usepackage{mathrsfs}%
\usepackage[title]{appendix}%
\usepackage{xcolor}%
\usepackage{textcomp}%
\usepackage{manyfoot}%
\usepackage{booktabs}%
\usepackage{algorithm}%
\usepackage{algorithmicx}%
\usepackage{algpseudocode}%
\usepackage{listings}
\usepackage{amsopn}
\usepackage{amstext}%

\theoremstyle{Thmstyleone}%

\theoremstyle{Thmstyletwo}%

\theoremstyle{Thmstylethree}%

\begin{document}

\title[Neural Stochastic Forecasting for Lateral Molecular Diffusion Simulations]{Neural Stochastic Forecasting for Lateral Molecular Diffusion Simulations}

\author[1,2]{\fnm{Jacob Ede} \sur{Levine}}\email{jlevine@cpp.edu}
\equalcont{}

\author[2]{\fnm{Mario} \sur{Hidalgo-Soria}}\email{mhidalgosoria@westernu.edu}
\equalcont{These authors contributed equally to this work.}

\author[2]{\fnm{Chen Yun} \sur{Wen}}\email{chenyun.wen@westernu.edu}

\author*[2]{\fnm{Yun Lyna} \sur{Luo}}\email{luoy@westernu.edu}
\author*[1]{\fnm{Sai Chandra} \sur{Kosaraju}}\email{skosaraju@cpp.edu}
\affil*[1]{\orgdiv{Department of Computer Science}, \orgname{California State Polytechnic University}, \orgaddress{\city{Pomona}, \postcode{91768}, \state{California}, \country{USA}}}

\affil[2]{\orgdiv{Department of Biotechnology and Pharmaceutical Sciences}, \orgname{Western University of Health Sciences}, \orgaddress{\city{Pomona}, \postcode{91766}, \state{California}, \country{USA}}}

\abstract{

Predicting the long-time diffusion of membrane proteins is challenging because molecular dynamics (MD) simulations are computationally expensive, while experimental single-particle tracking is often limited by finite observation times. Here, we introduce a neural stochastic (NS) model for forecasting the collective diffusion of membrane proteins from coarse-grained MD trajectories. The model combines a transformer-based representation of protein configurations, recurrent networks operating at multiple temporal scales, and a stochastic latent process with learned memory. A cross-attention decoder then generates correlated protein displacements while accounting for the instantaneous system geometry and periodic boundary conditions. We evaluate the model in two membrane systems: seven interacting PIEZO channels and a mixed system containing seven PIEZO and ten TREK channels. For the seven PIEZO system, the NS model reproduces the anomalous diffusion exponent measured in MD simulations $\alpha \sim 1$, substantially improving on a previously developed variational autoregressive model that exhibits super-diffusive behavior. The model also generates trajectories extending to 40,000 frames and captures the distinct diffusion characteristics of PIEZO and TREK populations in the mixed system. Comparisons of time-averaged mean-square displacements, Van Hove displacement distributions, and non-Gaussian parameters show that the model preserves key statistical features of the simulated dynamics. These results establish neural stochastic modeling as a promising approach for generating long trajectories of multi-protein membrane diffusion and extending the statistical analysis of molecular dynamics simulations.
}

\maketitle

\section{Introduction}\label{sec1}
Protein membrane diffusion is a widely studied phenomenon within the realm of cellular transport ~\cite{javanainen2012}. Its biological significance is crucial for understanding vital cellular activities like cell signaling ~\cite{cho2005}, and membrane domain organization ~\cite{owen2009}. Further, the study of protein membrane diffusion has been applied to identify disease mechanisms in the cell ~\cite{alenghat2013} and drug targeting ~\cite{international2010}.

Traditionally single particle tracking (SPT) experiments are implemented to obtain trajectories of individual proteins in the plasma membrane ~\cite{saxton1997}. SPT techniques allow to reconstruct individual membrane protein time series with nano-meter scale localization precision and mili-second temporal resolution ~\cite{barkai2012}. Despite its high spatiotemporal resolution, fluorescence based SPT is generally limited by photobleaching, blinking, and molecular loss from the observation region, resulting in individual trajectories that typically span tenth of seconds ~\cite{ly2025}, although specialized labeling and imaging  can extend observation times to minutes ~\cite{ly2025}.

Due to these limitations exhibited by SPT experiments,  \textit{in silico} methods like molecular dynamics (MD) simulations, are commonly employed to investigate protein membrane diffusion and other cellular dynamics over longer time scales. However MD simulations are highly expensive in terms of computational power since they represent solving numerically all the equations of motion for the atoms or groups of atoms forming the system in question.

Recently in ~\cite{thai2026} we introduced a collection of variational generative models to forecast protein membrane diffusion. Showing that a stochastic latent representation  in a variational auto-regressive (VAR) architecture was good enough to capture short-time multi-protein membrane diffusion. Nevertheless, these models were not able to replicate/forecast long time dynamics.

In this work we introduce a physics informed neural stochastic model which let us to forecast long time transport of different  protein membrane systems.  Our model is tested in a multi-protein system with 7 PIEZO channels diffusing in a membrane bilayer \cite{thai2026,wen2025}. Then our model is implemented to predict the diffusion in a mixed multi-protien system where 7 PIEZO channels interact and diffuse altogether with 10 TREK channels ~\cite{glogowska2021}. We show how the inclusion of a physics informed model within a neural stochastic framework is able to describe accurate long-term diffusion regimes and forecast MD trajectories beyond simulated length.

\section{Methods}

The model describes the motion of membrane proteins with a latent stochastic process coupled to neural networks that encode the protein configuration and its history.
The neural networks determine the mean of the latent process and convert each sample of the latent state into displacements of the proteins.
The model is trained by matching statistics of physical observables between generated trajectories and coarse-grained molecular dynamics (MD) trajectories.
Section~\ref{subsec:methods-overview} formulates the model, and Section~\ref{subsec:methods-md} describes the MD datasets.
Sections~\ref{subsec:methods-aug} to \ref{subsec:methods-decoder} describe the training data and the network components, and Sections~\ref{subsec:methods-loss} to \ref{subsec:methods-metrics} describe the training loss, the optimization, the evaluation, the model selection and the metrics.

\subsection{Neural stochastic differential equation}%
\label{subsec:methods-overview}

Each MD trajectory is a sequence of $T$ frames, $\{\mathbf{X}^{(0)}, \ldots, \mathbf{X}^{(T-1)}\}$, where the superscript $t$ indexes the frame.
Each frame $\mathbf{X}^{(t)} \in \mathbb{R}^{N \times 3}$ contains the three-dimensional positions of the $N$ proteins of the system.
Row $i$ of a frame, $\mathbf{X}^{(t)}_i \in \mathbb{R}^3$, is the position of protein $i = 1, \ldots, N$ in \AA.
The position of each protein is defined for each system in Section~\ref{subsec:methods-md}.

The model represents the effect of the unrecorded degrees of freedom on the protein motion through a latent stochastic state.
This state is propagated forward in time under a learned model of its fluctuations.
The recorded positions are treated as a stochastic and, in general, non-Markovian process, in which the future motion of a protein depends on its past trajectory and not only on its current configuration.
The MD trajectories are generated by numerical integration of the equations of motion of a coarse-grained force field whose potential couples the proteins to the surrounding lipids.
The reduced trajectories retain one position per protein and do not contain the lipid coordinates, the internal coordinates of the proteins or the forces acting on the proteins.

Each trajectory generated by the model, referred to as a rollout, is initialized from a single MD configuration.
After the initial configuration, every input to the model is the output of the model at the previous step, during both training and inference.
Unlike a next-frame predictor, the model is not conditioned on a window of observed frames.

Each step of a rollout advances the configuration by one frame through five modules applied in sequence.
The modules are a transformer encoder, a multi-rate bank of long short-term memory (LSTM) cells, a latent transition head, a latent stochastic integrator and a cross-transformer decoder.
Each module is summarized in the following paragraphs and described in detail in Sections~\ref{subsec:methods-spatial} to \ref{subsec:methods-decoder}.

The transformer encoder (Section~\ref{subsec:methods-spatial}) converts the current frame into a geometry vector, which is the input of the LSTM bank and the decoder.
The input of the encoder is one vector per protein, which encodes the absolute position of the protein at the initial frame and the most recent displacement of the protein at every later step.
A self-attention layer exchanges information among the proteins and produces one feature vector per protein.
The geometry vector consists of these feature vectors and the distances between all pairs of proteins.

The multi-rate LSTM bank (Section~\ref{subsec:methods-temporal}) summarizes the history of the geometry vector over multiple time scales.
The bank consists of LSTM cells that are updated at different temporal strides, from every frame to once every $300$ frames.
A cell with stride $\lambda$ is updated once every $\lambda$ frames and holds its state between updates.

The latent transition head (Section~\ref{subsec:methods-transition}) maps the output of the LSTM bank to the mean of a latent state.
The latent state has six components per protein, and these components have no prescribed physical meaning.
Each component of the mean depends on the LSTM states of all proteins.

The latent stochastic integrator (Section~\ref{subsec:methods-sde}) advances the latent state by one step of a discrete Ornstein-Uhlenbeck (OU) process.
At each step, the latent state relaxes toward the mean produced by the transition head and receives a Gaussian random increment.
The process has two learned parameters, a memory coefficient and a noise amplitude.
The memory coefficient sets the fraction of the previous deviation from the mean that is retained at each step, and the noise amplitude sets the magnitude of the random increment.

The cross-transformer decoder (Section~\ref{subsec:methods-decoder}) converts the latent state into one displacement per protein.
In its attention layer, the attention weights are computed from the geometry vector, and the information exchanged between proteins is taken from their latent states.
Through this layer, each protein combines its own latent state with the latent states of the other proteins.
For each protein, the decoder outputs the mean and a full $3 \times 3$ covariance matrix of a Gaussian distribution of the three-dimensional displacement, which includes correlations between the Cartesian components.
One displacement is sampled from this distribution for each protein, and the protein position is advanced by this displacement and wrapped into the periodic simulation box.

Temporal correlations over a rollout are represented by the LSTM bank and the latent process.
The LSTM bank represents temporal correlations through the mean of the latent state.
The latent process represents temporal correlations through its memory coefficient, to an extent that increases with the value of this coefficient.
After the initial step, the geometry vector is computed only from the most recent displacement of each protein and the current pair distances.
The decoder retains no information between steps and determines only the instantaneous mean and covariance of the displacement.

The model is trained autoregressively, in the same mode used at inference, without teacher forcing.
Teacher forcing replaces the predicted frame with the measured frame as the input at each training step \cite{williams1989}.

The loss function compares statistics of physical observables computed over a rollout with the corresponding MD trajectory.
Apart from the initial configuration, the MD trajectory enters training only through this loss.
All loss terms except one compare distributions or averages rather than individual frames.
The single exception, defined in Section~\ref{subsec:methods-loss}, is a pointwise comparison of the products of single-frame displacements used to compute the velocity autocorrelation function.

Training rollouts and evaluation rollouts differ only in their length, in the use of dropout and in the arithmetic precision (Sections~\ref{subsec:methods-schedule} and \ref{subsec:methods-stability}).
A training rollout spans $2{,}999$ steps, and an evaluation rollout spans $40{,}000$ steps.
Dropout \cite{srivastava2014dropout}, which sets randomly selected activations to zero, is applied during training and disabled during evaluation.
Training uses mixed-precision arithmetic \cite{micikevicius2018mixed} with the bfloat16 format \cite{kalamkar2019bfloat16}, a 16-bit floating-point format with the exponent range of single precision, whereas evaluation uses single precision.

The evaluation in Section~\ref{subsec:methods-stability} compares statistics of evaluation rollouts with those of the MD trajectories at lag times of up to $5{,}000$ frames.
Model selection and reporting use this evaluation (Section~\ref{subsec:methods-selection}).

\subsection{Molecular dynamics datasets}\label{subsec:methods-md}

We use two coarse-grained systems, PIEZO and PIEZO-TREK, summarized in Table~\ref{tab:methods-datasets}.
Both systems were simulated with LAMMPS \cite{thompson2022lammps}, and each trajectory is reduced to one position per protein per frame.
The PIEZO system contains seven PIEZO channels, and the PIEZO-TREK system contains seven PIEZO channels and ten TREK channels.
The same architecture and the same training procedure are applied to both systems.
The systems differ in the number of protein species and in the loss weights of Section~\ref{subsec:methods-loss}.

The species of a protein determines which parameters of the model it shares with other proteins.
Each species has its own input projection in the transformer encoder, its own LSTM bank, its own latent dynamics parameters, and its own output heads in the decoder (Sections~\ref{subsec:methods-spatial} to \ref{subsec:methods-decoder}).
Proteins of the same species share these parameters, and the remaining parameters are common to all proteins.

\begin{table}[ht]
\caption{Coarse-grained systems. Frames are counted after the removal of duplicated frames described in the text. $L_{\text{raw}}$ is the raw in-plane box edge, $\sigma_\Delta$ the displacement scale defined after Equation~\ref{eq:methods:minimage-raw}, and $L = L_{\text{raw}} / \sigma_\Delta$ the dimensionless normalized box edge of Equation~\ref{eq:methods:normalize}.}
\label{tab:methods-datasets}
\begin{tabular}{@{}lll@{}}
\toprule
 & PIEZO & PIEZO-TREK \\
\midrule
Proteins $N$                       & 7                & 17 \\
Composition                        & 7 PIEZO          & 7 PIEZO, 10 TREK \\
Species                            & 1                & 2 \\
Relabeling symmetry                & all permutations & within each species \\
Replicas used                      & 2                & 2 \\
Frames, replica 1                  & 12{,}578         & 6{,}001 \\
Frames, replica 2                  & 7{,}103          & 9{,}002 \\
Training frames                    & \multicolumn{2}{c}{first 3{,}000 frames of replica 1} \\
In-plane box edge $L_{\text{raw}}$ & 3440 \AA    & 3400 \AA \\
Displacement scale $\sigma_\Delta$ & 4.018202 \AA     & 6.298365 \AA \\
Normalized box $L$                 & 856.196          & 539.823 \\
Protein pairs                      & 21               & 136 \\
\bottomrule
\end{tabular}
\end{table}

The PIEZO system contains seven copies of a coarse-grained elastic-network model \cite{wen2025} of the mechanosensitive channel PIEZO \cite{coste2010piezo}, diffusing in a large tension-free bilayer.
The model is a three-fold symmetric trimer of $1869$ beads built on the cryo-electron microscopy structure of mouse PIEZO2 (PDB 6KG7) \cite{wang2019piezo2}.
At this resolution, the model does not distinguish PIEZO1 from PIEZO2 \cite{wen2025}.

The position of each PIEZO channel is the position of its central marker bead.
Each channel carries four marker beads, one at the center and three on the arms.
The three arm markers, whose positions fluctuate under constant pressure, are not used.
The central markers are unwrapped across the periodic boundary, the initial frames of each replica are removed, and the production segments of each replica are joined into one trajectory with one unwrapped position per protein per frame.

Two independent replicas of the PIEZO system are used.
Replica 1 has $12{,}632$ frames, or $12{,}578$ frames after the removal of duplicated frames described below, and replica 2 has $7{,}103$ frames.
A single Frames represents $10{,}000$ MD steps, corresponding to $0.5$~ns per frame with an integration time step of $50$~fs.

The seven PIEZO channels of the PIEZO system are identical copies of one protein and are treated as a single species.
The statistics of the system are then invariant under any permutation of the seven protein indices.

Both PIEZO replicas are wrapped at the in-plane box edge of replica 1, $3440$~\AA.
The in-plane box edge of replica 2 is $3439$~\AA.
The PIEZO positions are unwrapped before they are wrapped at the common edge.

The PIEZO-TREK system contains seven PIEZO channels and ten TREK channels diffusing together in one bilayer with an in-plane edge of $3400$~\AA.
TREK channels are two-pore-domain potassium channels of the TREK subfamily \cite{goldstein2005k2p}.
The system contains $17$ proteins and $136$ protein pairs.

The PIEZO and TREK channels of the PIEZO-TREK system are treated as two species.
The statistics of the system are invariant under permutations of the seven PIEZO indices and under permutations of the ten TREK indices, and they are not invariant under an exchange of a PIEZO channel with a TREK channel.

Each protein of the PIEZO-TREK system is represented by one position per frame.
The simulation provides four three-dimensional points for each PIEZO channel and one point for each TREK channel.
The position of a PIEZO channel is the first of its four points, and the position of a TREK channel is its single point.
These positions are wrapped and scaled with Equation~\ref{eq:methods:normalize} without further processing.

Replica 1 of the PIEZO-TREK system has $11{,}001$ frames total in the trajectory.

Replica 2 of the PIEZO-TREK system is converted to the box of replica 1.
This replica has $9{,}002$ frames total in the trajectory
It was simulated in a box with an in-plane edge of $3296.94$~\AA\ and an edge of $600$~\AA\ along the membrane normal.
Its minimum-image steps are computed in this box, accumulated from its first frame, expressed in the length unit of replica 1 and wrapped into the box of replica 1, without rescaling of the coordinates.
This conversion preserves every single-protein displacement exactly and does not preserve the separation of two proteins across the box boundary.

Replica 2 of the PIEZO-TREK system is used only in the evaluation reference of Section~\ref{subsec:methods-stability} and in the allowance of Section~\ref{subsec:methods-allowance}.
It supplies neither training frames nor initial conditions.

The coordinates are wrapped into the periodic simulation box and divided by a fixed displacement scale $\sigma_\Delta$.
The scale is computed from the single-step minimum-image displacements
\begin{equation}
\Delta\mathbf{X}^{(t)}_i = \mathbf{X}^{(t+1)}_i - \mathbf{X}^{(t)}_i
- L_{\text{raw}}\,\operatorname{round}\!\left(
\frac{\mathbf{X}^{(t+1)}_i - \mathbf{X}^{(t)}_i}{L_{\text{raw}}}\right)
\label{eq:methods:minimage-raw}
\end{equation}
which are evaluated component-wise for every protein and every pair of consecutive frames.
Here $\operatorname{round}(\cdot)$ denotes rounding to the nearest integer, and $L_{\text{raw}}$ is the raw in-plane box edge in \AA.
The subtraction replaces each step by the shortest step between periodic images and removes the jumps of wrapped coordinates by integer multiples of the box edge.
The displacement scale $\sigma_\Delta$, in \AA, is the standard deviation about the mean of all three components of $\Delta\mathbf{X}^{(t)}_i$, pooled over all proteins and over the frames specified below for each system.

The normalized coordinates and the normalized box edge are
\begin{equation}
\mathbf{x}^{(t)}_i = \bigl(\mathbf{X}^{(t)}_i \bmod L_{\text{raw}}\bigr) /
\sigma_\Delta,
\qquad
L = L_{\text{raw}} / \sigma_\Delta ,
\label{eq:methods:normalize}
\end{equation}
where $\mathbf{x}^{(t)}_i \in [0, L)^3$ is the dimensionless wrapped position of protein $i$ and $L$ is the dimensionless box edge used in all subsequent equations.
The modulo operation $y \bmod L_{\text{raw}} = y - L_{\text{raw}}\lfloor y / L_{\text{raw}} \rfloor \in [0, L_{\text{raw}})$, with $\lfloor\cdot\rfloor$ the largest integer not greater than its argument, is applied component-wise.
The normalized single-step displacements are of order unity.

In both systems, the membrane-normal component is wrapped at the in-plane edge $L_{\text{raw}}$ in Equations~\ref{eq:methods:minimage-raw} and \ref{eq:methods:normalize}.
The recorded box edge along the membrane normal is $2000$~\AA\ for the PIEZO system and $600$~\AA\ for replica 2 of the PIEZO-TREK system, and it is not recorded for replica 1 of the PIEZO-TREK system.
All membrane-normal steps in these trajectories are shorter than $L_{\text{raw}}/2$, and Equation~\ref{eq:methods:minimage-raw} leaves them unchanged.

The displacement scale is computed once for each system and applied to all replicas of that system.
For the PIEZO system, $\sigma_\Delta$ is computed over both replicas pooled.
For the PIEZO-TREK system, $\sigma_\Delta$ is computed over the $11{,}003$ frames of replica 1, and replica 2 is expressed in the same unit.
In both systems, $\sigma_\Delta$ is computed over all species jointly rather than per species.
Physical units are recovered by multiplication of a coordinate or a displacement by $\sigma_\Delta$.

Within the model, positions are wrapped and displacements are computed in the normalized box of edge $L$.
All three normalized coordinates are wrapped at the in-plane edge $L$ (Equation~\ref{eq:methods:state-update}).
Displacements are computed with the minimum-image map \cite{allen2017computer} $\operatorname{mi}(\mathbf{y}) = \mathbf{y} - L\,\operatorname{round}(\mathbf{y}/L)$, which is applied component-wise to a difference $\mathbf{y}$ of normalized positions.

Lateral statistics are computed on unwrapped lateral positions $\mathbf{r}^{(t)}_i \in \mathbb{R}^2$, which are accumulated from minimum-image steps,
\begin{equation}
\mathbf{r}^{(t_0)}_i = \mathbf{x}^{(t_0)}_{i,\mathrm{xy}},
\qquad
\mathbf{r}^{(t+1)}_i = \mathbf{r}^{(t)}_i +
\operatorname{mi}\!\bigl(\mathbf{x}^{(t+1)}_{i,\mathrm{xy}} -
\mathbf{x}^{(t)}_{i,\mathrm{xy}}\bigr),
\label{eq:methods:unwrap}
\end{equation}
where the subscript $\mathrm{xy}$ selects the two lateral components and $t_0$ is the first frame of the trajectory.
Equation~\ref{eq:methods:unwrap} recovers the continuous lateral path when every single-frame lateral displacement is shorter than $L/2$.
Every lateral step of the training data satisfies this condition in both systems.

\subsection{Training data, symmetry augmentation and sampling}%
\label{subsec:methods-aug}

Each model is trained on the first $3{,}000$ frames of replica 1 of its system.
For the PIEZO system, these frames span $1.5$~\textmu s.
The training views are generated from these frames by three symmetry operations, namely in-plane rotations and reflections, in-plane translations and time reversal.
These operations produce equivalent views of the same frames and add no information beyond the MD data.

The dihedral group $D_4$ of the square provides eight views of each trajectory.
The group consists of the four in-plane rotations by multiples of $90^\circ$, with and without reflection.
The in-plane box of both systems is square and periodic.
The rotations and reflections act on the lateral coordinates about the center of the box, the result is wrapped into the box, and the membrane-normal coordinate is unchanged.

Each $D_4$ view is shifted by $1{,}500$ rigid in-plane translations, which are exact symmetries of the periodic box.
The translations consist of the identity and $1{,}499$ shifts drawn uniformly from $[0, L)^2$ with a fixed seed, and the lateral coordinates are wrapped after each shift.
The same $1{,}500$ shifts are applied to every element of $D_4$.
The translations change the absolute positions that the transformer encoder receives at the initial frame (Section~\ref{subsec:methods-spatial}).

Each view is also used with its frames in reversed order.
Every statistic of the evaluation in Section~\ref{subsec:methods-stability} is even in the displacement and takes the same value on a trajectory and on its time reversal.
The cohesion term of the loss (Section~\ref{subsec:methods-loss}) matches signed increments of pair distances, which change sign under time reversal.
With both directions in the training set, this term matches a symmetrized distribution of these increments.
A model trained on both directions learns a time-symmetrized transition, whether or not the MD dynamics are reversible.

The training set of each system consists of $8 \times 1{,}500 \times 2 = 24{,}000$ views of the same $3{,}000$ frames, half in forward order and half in reversed order.
Particle relabeling, which is a symmetry within each species (Table~\ref{tab:methods-datasets}), is not used for augmentation.

Each training item is one of the $24{,}000$ views, selected uniformly at random, and uses all $3{,}000$ frames of the view as one window.
Within a window, frame $0$ is the initial condition and frames $1$ to $2{,}999$ are the targets of Section~\ref{subsec:methods-loss}.
An epoch draws $960$ windows in batches of $64$, which gives $15$ optimizer steps per epoch.

No validation split is held out.
Model selection uses the long-horizon evaluation of Section~\ref{subsec:methods-stability}.

\subsection{Notation and network components}\label{subsec:methods-notation}

Vectors are columns set in bold, and matrices are set in bold capitals.
The symbol $\odot$ denotes element-wise multiplication, and $[\cdot\,, \cdot]$ denotes the concatenation of vectors into a longer vector.
The Euclidean norm is denoted $\|\cdot\|$, and $|\cdot|$ denotes the absolute value of a number or the number of elements of a set.
For a vector, $\operatorname{mean}(\cdot)$ and $\operatorname{Var}(\cdot)$ denote the mean and the variance of its entries.
The brackets $\langle\cdot\rangle$ denote an average over the set named in the subscript or in the text.
The function $\operatorname{atan2}(y_1, y_2)$ returns the angle in $(-\pi, \pi]$ whose tangent is $y_1/y_2$, with the quadrant determined by the signs of both arguments.
Every use of $\operatorname{atan2}$ in the following subsections has a non-negative first argument, and the resulting angles lie in $[0,\pi]$.
A softmax over an index set normalizes the exponentials of its arguments to a sum of one over that set.

Upright $\mathrm{x}$, $\mathrm{y}$ and $\mathrm{z}$ denote the coordinate axes, and $\mathrm{z}$ is the membrane normal.
Each protein $i$ belongs to a species $s(i)$, $\mathcal{P}_s$ is the set of proteins of species $s$, and $\mathcal{S}$ is the set of species of a system.

Four element-wise nonlinear functions are used in the model.
Each function maps a real number $y$ to a real number and is applied independently to every entry of its argument,
\begin{align}
\operatorname{sigmoid}(y) &= \frac{1}{1 + \exp(-y)} \in (0,1),
&\quad
\operatorname{softplus}(y) &= \ln\!\bigl(1 + \exp(y)\bigr) \in (0,\infty),
\label{eq:methods:sigmoid-softplus}\\[2pt]
\operatorname{SiLU}(y) &= y\,\operatorname{sigmoid}(y),
&\quad
\operatorname{GELU}(y) &= y\,\Phi(y),
\label{eq:methods:silu-gelu}
\end{align}
where $\Phi$ is the cumulative distribution function of the standard normal distribution.
The sigmoid maps any real number into the open unit interval and is used for quantities that represent fractions or gates.
The softplus is a smooth approximation of $\max(0, y)$ and is used for quantities that must be strictly positive, such as standard deviations.
The SiLU \cite{hendrycks2016gelu,elfwing2018silu} and GELU \cite{hendrycks2016gelu} functions are smooth activation functions that pass large positive arguments almost unchanged and suppress negative arguments.
These two functions serve as the activation functions of the model in place of rectified linear units.

Layer normalization \cite{ba2016layernorm} rescales a vector to zero mean and unit variance across its components and applies a learned affine transformation,
\begin{equation}
\operatorname{LN}(\mathbf{y}) = \boldsymbol{\gamma} \odot
\frac{\mathbf{y} - \operatorname{mean}(\mathbf{y})}{
\sqrt{\operatorname{Var}(\mathbf{y}) + 10^{-5}}} + \boldsymbol{\beta},
\label{eq:methods:layernorm}
\end{equation}
in which $\mathbf{y}$ is the input vector, $\boldsymbol{\gamma}$ and $\boldsymbol{\beta}$ are learned vectors of the same length, and the constant $10^{-5}$ prevents division by zero.
Each layer normalization in the model has its own $\boldsymbol{\gamma}$ and $\boldsymbol{\beta}$ and is identified by a subscript.

Every learned weight matrix is denoted $\mathbf{W}$ and every learned bias $\mathbf{b}$, each with a superscript that identifies its layer.
All parameters without a stated initialization use the default initialization of PyTorch.

Spectral normalization \cite{miyato2018spectral} is applied to the first linear layer of the transition head, of the value map of the decoder and of the mean and covariance heads of the decoder.
For a weight matrix $\mathbf{W}$ with largest singular value $\|\mathbf{W}\|_2$, the spectrally normalized weight is
\begin{equation}
\operatorname{sn}(\mathbf{W}) = \mathbf{W} / \|\mathbf{W}\|_2,
\label{eq:methods:spectral-norm}
\end{equation}
which satisfies $\|\operatorname{sn}(\mathbf{W})\,\mathbf{y}\| \le \|\mathbf{y}\|$ for every $\mathbf{y}$.

The divisor $\|\mathbf{W}\|_2$ is estimated by power iteration, which repeatedly multiplies a vector by $\mathbf{W}^{\top}\mathbf{W}$ and converges to the leading singular vector.
The estimate advances by one iteration at each evaluation of the layer during training, which occurs at every rollout step, and is held fixed during evaluation.
Only the first linear layer of each path is constrained, and the composite map is not constrained to be non-expansive.

\subsection{Transformer encoder}\label{subsec:methods-spatial}

One attention head combines value vectors with weights computed from scaled dot products of queries and keys \cite{vaswani2017attention}.
For matrices $\mathbf{Q}$, $\mathbf{K}$ and $\mathbf{V}$ whose rows are the query, key and value projections of a set of items, the head computes
\begin{equation}
\operatorname{Attn}(\mathbf{Q}, \mathbf{K}, \mathbf{V}) =
\operatorname{softmax}\!\left(
\frac{\mathbf{Q}\mathbf{K}^{\top}}{\sqrt{d}}\right)\mathbf{V},
\label{eq:methods:attn}
\end{equation}
where $d$ is the number of components of the query and key of one head and the softmax is taken along each row.
The weights that one item places on the other items are non-negative and sum to one, and each output row is a weighted average of value rows.

Several attention heads operate in parallel on separate learned projections, and their outputs are concatenated and mixed by a learned matrix.
An attention block adds its output to its input through a residual connection \cite{he2016resnet} and is followed by a feedforward network applied to each item separately.
In an attention layer, the vector that represents one item is called a token.

The transformer encoder is a single encoder layer of the original transformer \cite{vaswani2017attention}, with layer normalization after each residual connection, applied to the set of protein tokens without positional encoding.
In this form, the encoder is the set attention block of the Set Transformer \cite{lee2019settransformer}, which treats its tokens as an unordered set.
The encoder has no mask and no edge list, and every protein attends to every protein, including itself.
On the complete graph of proteins, the encoder is a form of graph attention \cite{velickovic2018gat}, with weights from the scaled dot products of Equation~\ref{eq:methods:attn} between node features instead of the additive score of the original graph attention network, and without restriction of the softmax to the neighbors of a node.

The attention weights of the encoder do not depend on the distances between proteins.
After the initial frame, the node features contain only the most recent step of each protein, and the attention weights depend on this motion and on the species of the proteins but not on their positions.
The spatial arrangement of the proteins enters the model only through the pairwise distances, which are appended after the attention block (Equation~\ref{eq:methods:pairdist}) and are used by the LSTM bank and the decoder.

The node input of protein $i$ at step $t$ is a nine-component vector,
\begin{equation}
\mathbf{f}^{(t)}_i =
\begin{cases}
\bigl[\sin\bigl(2\pi\mathbf{x}^{(0)}_i/L\bigr),\;
\cos\bigl(2\pi\mathbf{x}^{(0)}_i/L\bigr),\; \mathbf{0}_3\bigr], & t = 0,\\[4pt]
\bigl[\mathbf{0}_6,\;
\operatorname{mi}\bigl(\mathbf{x}^{(t)}_i - \mathbf{x}^{(t-1)}_i\bigr)\bigr], & t \ge 1,
\end{cases}
\label{eq:methods:node-input}
\end{equation}
in which the sine and cosine act on each of the three components, the three sines precede the three cosines, and $\mathbf{0}_3$ and $\mathbf{0}_6$ are vectors of three and six zeros.
At the initial frame, the node input encodes the absolute position of the protein through a sine-cosine embedding, which is continuous across the periodic boundary and differentiable everywhere.
At every later step, the node input contains only the minimum-image step of the protein.
Absolute position enters the encoder only at the initial frame, and no gradient passes through the sine and cosine of the positions predicted by the model.

Each node input is projected to a $63$-component node feature by an affine map shared by the proteins of one species, and the node features form the rows of the node matrix $\mathbf{H}^{(t)} \in \mathbb{R}^{N\times 63}$,
\begin{equation}
\mathbf{H}^{(t)}_{i,:} = \bigl(\mathbf{W}^{\mathrm{node}}_{s(i)}\,
\mathbf{f}^{(t)}_i + \mathbf{b}^{\mathrm{node}}_{s(i)}\bigr)^{\!\top},
\label{eq:methods:node-proj}
\end{equation}
where $\mathbf{H}^{(t)}_{i,:}$ is row $i$, and $\mathbf{W}^{\mathrm{node}}_{s}$ and $\mathbf{b}^{\mathrm{node}}_{s}$ are the projection parameters of species $s$.
The PIEZO system has one projection, and the PIEZO-TREK system has one projection per species.

The proteins exchange information in one attention block and one feedforward block, each with a residual connection followed by layer normalization,
\begin{align}
\hat{\mathbf{H}}^{(t)} &= \operatorname{LN}_{\mathrm{G}1}\!\Bigl(
\mathbf{H}^{(t)} + \operatorname{MHA}\bigl(\mathbf{H}^{(t)}\bigr)\Bigr),
\label{eq:methods:enc-attn}\\
\tilde{\mathbf{H}}^{(t)} &= \operatorname{LN}_{\mathrm{G}2}\!\Bigl(
\hat{\mathbf{H}}^{(t)} +
\operatorname{GELU}\bigl(\hat{\mathbf{H}}^{(t)}\mathbf{W}^{\mathrm{G}1\top}\bigr)
\mathbf{W}^{\mathrm{G}2\top}\Bigr),
\label{eq:methods:enc-ffn}
\end{align}
where layer normalization acts on each row.
The multi-head attention $\operatorname{MHA}$ applies Equation~\ref{eq:methods:attn} in each of $21$ heads, with query, key and value projections of $\mathbf{H}^{(t)}$ from $63$ to $3$ components per head ($d = 3$), and concatenates the $21$ head outputs and multiplies them by a learned $63 \times 63$ matrix.
The feedforward weights $\mathbf{W}^{\mathrm{G}1}$ and $\mathbf{W}^{\mathrm{G}2}$ map $63$ components to $126$ and back, and biases are omitted from Equations~\ref{eq:methods:enc-attn} and \ref{eq:methods:enc-ffn}.
The transpose of row $i$ of $\tilde{\mathbf{H}}^{(t)}$, denoted $\mathbf{p}^{(t)}_i \in \mathbb{R}^{63}$, is the protein block of protein $i$.

The geometry vector combines the protein blocks with the minimum-image distances between all protein pairs.
The encoder computes the $N(N-1)/2$ three-dimensional pair distances
\begin{equation}
\rho^{(t)}_{ij} = \sqrt{\bigl\|\operatorname{mi}\bigl(\mathbf{x}^{(t)}_i -
\mathbf{x}^{(t)}_j\bigr)\bigr\|^2 + 10^{-8}}\;\Big/\,L,
\qquad 1 \le i < j \le N,
\label{eq:methods:pairdist}
\end{equation}
and collects them in the vector $\boldsymbol{\rho}^{(t)} \in \mathbb{R}^{N(N-1)/2}$.
The geometry vector is the concatenation of the protein blocks and this vector,
\begin{equation}
\mathbf{e}^{(t)} = \bigl[\mathbf{p}^{(t)}_1, \ldots, \mathbf{p}^{(t)}_N,\;
\boldsymbol{\rho}^{(t)}\bigr] \in \mathbb{R}^{63N + N(N-1)/2}.
\label{eq:methods:geometry-embed}
\end{equation}

The pair distances are expressed as fractions of the box edge and include the membrane-normal component.
The division by $L$ in Equation~\ref{eq:methods:pairdist} sets this scale, and the constant $10^{-8}$ keeps the square root differentiable at zero separation.
The membrane-normal coordinate, which no loss term constrains (Section~\ref{subsec:methods-loss}), enters the model through these distances and through the steps of Equation~\ref{eq:methods:node-input}.

The geometry vector is the complete description of one frame that the LSTM bank and the decoder receive.
It has $462$ components for the PIEZO system and $1207$ components for the PIEZO-TREK system (Table~\ref{tab:methods-dims}).
Protein $i$ occupies components $63(i-1)+1$ through $63i$ of $\mathbf{e}^{(t)}$, and the $N(N-1)/2$ distances occupy the final components.
The LSTM bank of Section~\ref{subsec:methods-temporal} uses this layout to route the features of each protein to the bank of its species.

\subsection{Multi-rate LSTM bank}\label{subsec:methods-temporal}

Each cell of the bank is an LSTM cell \cite{hochreiter1997lstm} with a forget gate \cite{gers2000forget}.
For an input $\tilde{\mathbf{e}}$, a previous hidden state $\mathbf{h}_{\mathrm{prev}}$ and a previous cell state $\mathbf{c}_{\mathrm{prev}}$, the cell computes three gates and a candidate content,
\begin{align}
\mathbf{g}^{\mathrm{in}} &= \operatorname{sigmoid}\bigl(
\mathbf{W}^{\mathrm{in}}[\tilde{\mathbf{e}}, \mathbf{h}_{\mathrm{prev}}]
+ \mathbf{b}^{\mathrm{in}}\bigr),
&\quad
\mathbf{g}^{\mathrm{fg}} &= \operatorname{sigmoid}\bigl(
\mathbf{W}^{\mathrm{fg}}[\tilde{\mathbf{e}}, \mathbf{h}_{\mathrm{prev}}]
+ \mathbf{b}^{\mathrm{fg}}\bigr),
\label{eq:methods:lstm-if}\\
\mathbf{g}^{\mathrm{out}} &= \operatorname{sigmoid}\bigl(
\mathbf{W}^{\mathrm{out}}[\tilde{\mathbf{e}}, \mathbf{h}_{\mathrm{prev}}]
+ \mathbf{b}^{\mathrm{out}}\bigr),
&\quad
\tilde{\mathbf{c}} &= \tanh\bigl(
\mathbf{W}^{\mathrm{cand}}[\tilde{\mathbf{e}}, \mathbf{h}_{\mathrm{prev}}]
+ \mathbf{b}^{\mathrm{cand}}\bigr),
\label{eq:methods:lstm-og}
\end{align}
and updates its two states,
\begin{equation}
\mathbf{c} = \mathbf{g}^{\mathrm{fg}} \odot \mathbf{c}_{\mathrm{prev}} +
\mathbf{g}^{\mathrm{in}} \odot \tilde{\mathbf{c}},
\qquad
\mathbf{h} = \mathbf{g}^{\mathrm{out}} \odot \tanh(\mathbf{c}).
\label{eq:methods:lstm-state}
\end{equation}

The three gates take values between zero and one and act as soft switches.
The forget gate $\mathbf{g}^{\mathrm{fg}}$ sets the fraction of the previous memory $\mathbf{c}_{\mathrm{prev}}$ that is retained, the input gate $\mathbf{g}^{\mathrm{in}}$ sets the fraction of the candidate content $\tilde{\mathbf{c}}$ that is written, and the output gate $\mathbf{g}^{\mathrm{out}}$ sets the fraction of the memory that is exposed as the hidden state $\mathbf{h}$.
The cell is written compactly as $\operatorname{LSTM}(\tilde{\mathbf{e}}, \mathbf{h}_{\mathrm{prev}}, \mathbf{c}_{\mathrm{prev}}) \mapsto (\mathbf{h}, \mathbf{c})$.

The LSTM bank consists of seven single-layer LSTM cells with $64$-component states, one for each stride $\lambda \in \{1, 3, 10, 30, 100, 200, 300\}$.
The cells run at different sampling rates, a design related to recurrent networks whose modules operate at separate clock rates \cite{koutnik2014clockwork}.
There is one bank per species, and all proteins of a species share it.
Every entry of $\mathbf{b}^{\mathrm{fg}}$ is initialized to two, and each cell starts training with its forget gate open.

The input to the bank of protein $i$ consists of the protein block of that protein and the pairwise distances,
\begin{equation}
\tilde{\mathbf{e}}^{(t)}_i = \operatorname{LN}_{s(i)}\bigl(
[\,\mathbf{p}^{(t)}_i,\; \boldsymbol{\rho}^{(t)}\,]\bigr)
\in \mathbb{R}^{63 + N(N-1)/2}.
\label{eq:methods:bank-input}
\end{equation}
The pairwise distances are common to all proteins, and the normalization statistics are computed separately for each protein.
The normalization $\operatorname{LN}_{s}$ computes its mean and variance over the vector of each protein, and its learned scale and shift are shared within species $s$, as are the cell weights.

Cell $\lambda$ of the bank of species $s(i)$ is updated only at steps that are multiples of its stride and otherwise keeps its states unchanged,
\begin{equation}
\bigl(\mathbf{h}^{(t)}_{i,\lambda},\, \mathbf{c}^{(t)}_{i,\lambda}\bigr) =
\begin{cases}
\operatorname{LSTM}_{s(i),\lambda}\bigl(\tilde{\mathbf{e}}^{(t)}_i,\,
\mathbf{h}^{(t-1)}_{i,\lambda},\, \mathbf{c}^{(t-1)}_{i,\lambda}\bigr)
& \text{if } t \bmod \lambda = 0, \\[4pt]
\bigl(\mathbf{h}^{(t-1)}_{i,\lambda},\, \mathbf{c}^{(t-1)}_{i,\lambda}\bigr)
& \text{otherwise,}
\end{cases}
\label{eq:methods:lag-update}
\end{equation}
where $\mathbf{h}^{(t)}_{i,\lambda}$ and $\mathbf{c}^{(t)}_{i,\lambda}$ are the states of protein $i$ in the cell of stride $\lambda$, $\operatorname{LSTM}_{s,\lambda}$ is the cell of Equations~\ref{eq:methods:lstm-if} to \ref{eq:methods:lstm-state} with the weights of species $s$ and stride $\lambda$, and both states start from zero, $\mathbf{h}^{(-1)}_{i,\lambda} = \mathbf{c}^{(-1)}_{i,\lambda} = \mathbf{0}$.
The step index $t$ counts from the first frame of a training window or from the initial condition of a rollout, and every cell is updated at $t = 0$.
The proteins of one species are processed together with the same weights.

The states of a cell are held, not reset, at skipped steps.
The cell of stride $1$ is updated at every step, and the cell of stride $300$ is updated once every $300$ steps and applies its forget gate once per $300$ frames.
After the initial frame, each update of a cell receives the single-frame step of Equation~\ref{eq:methods:node-input} and the current pair distances, and not the displacement accumulated over its stride.

The hidden states of all proteins and strides are concatenated in protein order and normalized,
\begin{equation}
\boldsymbol{\psi}^{(t)} = \operatorname{LN}_{\psi}\bigl(\bigl[
\mathbf{h}^{(t)}_{1,1}, \mathbf{h}^{(t)}_{1,3}, \ldots,
\mathbf{h}^{(t)}_{1,300}, \mathbf{h}^{(t)}_{2,1}, \ldots,
\mathbf{h}^{(t)}_{N,300}\bigr]\bigr) \in \mathbb{R}^{448N},
\label{eq:methods:readout}
\end{equation}
which gives $3136$ components for the PIEZO system and $7616$ components for the PIEZO-TREK system.

\subsection{Latent transition head}\label{subsec:methods-transition}

The latent state $\boldsymbol{\zeta}^{(t)} \in \mathbb{R}^{6N}$ has six components per protein, and components $6(i-1)+1$ through $6i$ form the slice $\boldsymbol{\zeta}^{(t)}_i \in \mathbb{R}^6$ of protein $i$.
The slices have no prescribed physical meaning.
The decoder uses slice $i$ as the value of protein $i$ (Section~\ref{subsec:methods-decoder}).

The transition head maps the bank readout to the mean $\boldsymbol{\mu}^{(t)} \in \mathbb{R}^{6N}$ toward which the latent state relaxes,
\begin{align}
\tilde{\boldsymbol{\mu}}^{(t)} &= \mathbf{W}^{\mu 2}\,\operatorname{SiLU}\!\bigl(
\operatorname{sn}\bigl(\mathbf{W}^{\mu 1}\bigr)\boldsymbol{\psi}^{(t)} + \mathbf{b}^{\mu 1}\bigr)
+ \mathbf{b}^{\mu 2},
\label{eq:methods:mu-head}\\
\boldsymbol{\mu}^{(t)} &= 10\,\tanh\!\bigl(\tilde{\boldsymbol{\mu}}^{(t)}/10\bigr),
\label{eq:methods:mu-clamp}
\end{align}
where $\mathbf{W}^{\mu 1}$ is a spectrally normalized layer of width $224N = \max(448N/2,\, 6N)$ and $\mathbf{W}^{\mu 2}$ is a linear layer onto the $6N$ components.
The hidden width is $1568$ for the PIEZO system and $3808$ for the PIEZO-TREK system.
The first layer of the head grows with the square of $N$ and contains most of the parameters of the model in both systems (Table~\ref{tab:methods-dims}).
The smooth clamp of Equation~\ref{eq:methods:mu-clamp} acts on each component, bounds the mean to the interval $(-10, 10)$ and is close to the identity for components much smaller than $10$ in magnitude.
The head outputs only the mean, and the noise of the latent process is generated by the stochastic integrator of Section~\ref{subsec:methods-sde}.

The transition head is fully connected across proteins, and each slice of $\boldsymbol{\mu}^{(t)}$ depends on the LSTM states of all proteins.
The head receives $\boldsymbol{\psi}^{(t)}$ with the proteins in a fixed order, and the model as a whole is not invariant to relabeling of the proteins within a species.
The transformer encoder, the LSTM banks and the decoder treat the proteins of a species identically.

\subsection{Latent stochastic integrator}\label{subsec:methods-sde}

The latent dynamics follow the framework of latent stochastic differential equations, in which observations are readouts of a hidden state that evolves under a drift and a noise parameterized by neural networks \cite{tzen2019neural,li2020scalable}.
The model is trained by matching rollout statistics (Section~\ref{subsec:methods-loss}) instead of the variational objectives of \cite{tzen2019neural,li2020scalable}.
The stochastic differential equation serves as the interpretation of the discrete update below.

The Ornstein-Uhlenbeck (OU) process \cite{uhlenbeck1930} is a mean-reverting process that relaxes exponentially toward its mean under white-noise forcing.
Observed at a fixed interval, the OU process is a first-order autoregressive process whose coefficient is the exponential of minus the ratio of the interval to the relaxation time \cite{gillespie1996ou}.

The latent state advances by one step of a discrete OU process toward the mean of Equation~\ref{eq:methods:mu-clamp}, and the slice of protein $i$ is updated as
\begin{equation}
\boldsymbol{\zeta}^{(t)}_i = \boldsymbol{\mu}^{(t)}_i +
a_{s(i)}\bigl(\boldsymbol{\zeta}^{(t-1)}_i - \boldsymbol{\mu}^{(t)}_i\bigr)
+ \kappa_{s(i)}\,\boldsymbol{\xi}^{(t)}_i,
\qquad
\boldsymbol{\xi}^{(t)}_i \sim \mathcal{N}(\mathbf{0}, \mathbf{I}_6),
\label{eq:methods:ou-randn}
\end{equation}
where $\boldsymbol{\mu}^{(t)}_i$ is the slice of protein $i$ of the mean, $a_s$ is the memory coefficient and $\kappa_s$ the noise amplitude of species $s$, and $\boldsymbol{\xi}^{(t)}_i$ is a vector of six independent standard normal numbers drawn at every step, with $\mathcal{N}(\mathbf{0}, \mathbf{I}_6)$ the standard normal distribution in six dimensions.
The update is computed in single precision, and during training with mixed precision its result is rounded to bfloat16 between steps.
The update starts from $\boldsymbol{\zeta}^{(-1)}_i = \mathbf{0}$, and the first state is $\boldsymbol{\zeta}^{(0)}_i = (1 - a_{s(i)})\boldsymbol{\mu}^{(0)}_i + \kappa_{s(i)}\boldsymbol{\xi}^{(0)}_i$.
No separate initialization of the latent state is used.

The memory coefficient and the noise amplitude are learned and are constrained to their admissible ranges by
\begin{equation}
a_s = \operatorname{sigmoid}(\tilde{a}_s) \in (0, 1),
\qquad
\kappa_s = \operatorname{softplus}(\tilde{\kappa}_s) \in (0, \infty),
\label{eq:methods:dynamics-params}
\end{equation}
where $\tilde{a}_s$ and $\tilde{\kappa}_s$ are the unconstrained parameters updated by the optimizer.
The parameters are initialized to $a_s = 0.8$ and $\kappa_s = 5.0$.
The PIEZO system has one pair of latent parameters, and the PIEZO-TREK system has one pair per species.
Both $\tilde{a}_s$ and $\tilde{\kappa}_s$ belong to a dedicated optimizer group (Section~\ref{subsec:methods-schedule}).

In Equation~\ref{eq:methods:ou-randn}, the latent state relaxes toward the target $\boldsymbol{\mu}^{(t)}_i$, retains the fraction $a_{s(i)}$ of its previous deviation from that target and receives a random increment of typical magnitude $\kappa_{s(i)}$.
The coefficient $a_s$ sets the memory of the process, and $\kappa_s$ sets the strength of its random forcing.
For $a_s$ close to zero, the latent state follows its mean, and for $a_s$ close to one, the latent state relaxes slowly toward its mean and accumulates its random increments.

For a fixed mean, Equation~\ref{eq:methods:ou-randn} is the exact one-frame transition of an OU process with a relaxation time of $-1/\ln a_s$ frames and a stationary standard deviation about the mean of $\kappa_s/(1 - a_s^2)^{1/2}$.
The mean of the model changes from step to step, and these two quantities describe the fluctuations of the latent state about its moving mean, not a stationary distribution.

The memory coefficient and the noise amplitude are parameterized directly, without a relaxation rate or an integration time step.
The model learns both parameters of the discrete process at the sampling interval of the data.

\subsection{Cross-transformer decoder}\label{subsec:methods-decoder}

Cross-attention is an attention operation in which the keys and values come from a different source than the queries \cite{vaswani2017attention}.
The original transformer refers to this operation as encoder-decoder attention.
The queries and keys determine which items exchange information, and the values carry the information that is exchanged.
A feedforward block, residual connections and layer normalization complete the layer.

The decoder is a cross transformer that maps the latent state to a displacement of each protein.
It consists of one attention block followed by one feedforward block and operates on one token per protein and one additional pair token.
The queries and keys are computed from the protein blocks of the transformer encoder, with one additional key computed from the pairwise distances, and the values are computed from the latent state.
In encoder-decoder attention, the keys and values come from the same source, and in this decoder the two sources are divided between the keys and the values.
The attention weights depend on the protein blocks and the pairwise distances and do not depend on the latent state.
A protein does not attend to itself, and its own latent slice is added directly to its token.

The protein token of protein $i$ is computed from its protein block, and the pair token is computed from the pairwise distances,
\begin{equation}
\mathbf{u}_i = \mathbf{W}^{\mathrm{u}}\operatorname{LN}_{\mathrm{u}}
\bigl(\mathbf{p}^{(t)}_i\bigr) + \mathbf{b}^{\mathrm{u}},
\qquad
\mathbf{u}_{\mathrm{pair}} = \mathbf{W}^{\mathrm{pair}}
\operatorname{LN}_{\mathrm{pair}}\bigl(\boldsymbol{\rho}^{(t)}\bigr)
+ \mathbf{b}^{\mathrm{pair}},
\label{eq:methods:dec-tokens}
\end{equation}
where both tokens have $256$ components, $\mathbf{W}^{\mathrm{u}}$ maps $63$ components to $256$, and $\mathbf{W}^{\mathrm{pair}}$ maps $N(N-1)/2$ components to $256$.
The step index $t$ is omitted from the quantities of the decoder.
No positional, index or species embedding is added, and the tokens of proteins of the same species are exchangeable.

Queries and keys are linear projections of the layer-normalized tokens,
\begin{equation}
\mathbf{q}_i = \mathbf{W}^{\mathrm{q}}\operatorname{LN}_{\mathrm{q}}(\mathbf{u}_i)
+ \mathbf{b}^{\mathrm{q}},
\qquad
\mathbf{k}_j = \mathbf{W}^{\mathrm{k}}\operatorname{LN}_{\mathrm{k}}(\mathbf{u}_j)
+ \mathbf{b}^{\mathrm{k}},
\quad j \in \{1, \ldots, N, \mathrm{pair}\},
\label{eq:methods:dec-qk}
\end{equation}
each with $256$ components split into four heads of $64$ components.
Only protein tokens issue queries, and the pair token serves only as a key.
In each head, the attention weight that protein $i$ places on token $j$ is
\begin{equation}
\omega_{ij} = \operatorname*{softmax}_{j \in \mathcal{U}_i}\!\left(
\frac{\mathbf{q}_i \cdot \mathbf{k}_j}{\sqrt{64}}\right),
\qquad
\mathcal{U}_i = \{1, \ldots, N, \mathrm{pair}\} \setminus \{i\},
\label{eq:methods:dec-attn}
\end{equation}
where the dot product runs over the $64$ components of the head and the softmax runs over the set $\mathcal{U}_i$, which excludes protein $i$.
The softmax is evaluated in single precision, and dropout at rate $0.252$ is applied to the attention weights during training.

After the initial frame, the attention weights among proteins do not depend on the separations between proteins.
The protein tokens then contain the recent steps and the species of the proteins and not their positions.
The key of the pair token depends on the pairwise distances and changes only the weight assigned to the pair token relative to the other proteins.

The value of protein $i$ is computed from its own latent slice, and the value of the pair token is zero,
\begin{equation}
\mathbf{v}_i = \operatorname{SiLU}\bigl(\operatorname{sn}\bigl(\mathbf{W}^{\mathrm{v}}\bigr)
\boldsymbol{\zeta}^{(t)}_i + \mathbf{b}^{\mathrm{v}}\bigr) \in \mathbb{R}^{256},
\qquad
\mathbf{v}_{\mathrm{pair}} = \mathbf{0},
\label{eq:methods:dec-values}
\end{equation}
where $\mathbf{W}^{\mathrm{v}}$ is a spectrally normalized map from six components to $256$.
Attention weight assigned to the pair token transmits no information from the other proteins.

The decoder representation of protein $i$ is formed from the attention output, the residual stream and a feedforward block,
\begin{align}
\mathbf{m}_i &= \Bigl[\textstyle\sum_{j \in \mathcal{U}_i} \omega_{ij}\,\mathbf{v}_j\Bigr]_{\text{4 heads}},
\label{eq:methods:dec-mix}\\
\mathbf{o}_i &= \mathbf{u}_i + \mathbf{v}_i + \mathbf{W}^{\mathrm{m}}\mathbf{m}_i
+ \mathbf{b}^{\mathrm{m}},
\label{eq:methods:dec-stream}\\
\hat{\mathbf{o}}_i &= \operatorname{LN}_{\mathrm{o}}\Bigl(\mathbf{o}_i +
\mathbf{W}^{\mathrm{f}2}\operatorname{GELU}\bigl(\mathbf{W}^{\mathrm{f}1}
\operatorname{LN}_{\mathrm{f}}(\mathbf{o}_i) + \mathbf{b}^{\mathrm{f}1}\bigr)
+ \mathbf{b}^{\mathrm{f}2}\Bigr),
\label{eq:methods:dec-ffn}
\end{align}
in which each head forms its weighted sum over its own $64$ value components and $[\cdot]_{\text{4 heads}}$ concatenates the four head outputs into $256$ components.
The residual stream of a token is its running representation, to which each block adds its output.
The matrix $\mathbf{W}^{\mathrm{m}}$ maps $256$ components to $256$, and $\mathbf{W}^{\mathrm{f}1}$ and $\mathbf{W}^{\mathrm{f}2}$ map $256$ components to $512$ and back.
During training, dropout at rate $0.252$ is also applied after the GELU and after $\mathbf{W}^{\mathrm{f}2}$, and it is omitted from the equations and inactive during evaluation.

Two output heads act on $\hat{\mathbf{o}}_i$, and each head consists of a spectrally normalized linear layer from $256$ to $128$ components, a SiLU and a final linear layer,
\begin{equation}
\begin{gathered}
\Delta\bar{\mathbf{x}}_i = \mathbf{W}^{\mathrm{dr}2}_{s(i)}\operatorname{SiLU}\bigl(
\operatorname{sn}\bigl(\mathbf{W}^{\mathrm{dr}1}_{s(i)}\bigr)\hat{\mathbf{o}}_i + \mathbf{b}^{\mathrm{dr}1}_{s(i)}\bigr)
+ \mathbf{b}^{\mathrm{dr}2}_{s(i)} \in \mathbb{R}^3,
\\
\boldsymbol{\varpi}_i = \mathbf{W}^{\mathrm{cov}2}_{s(i)}\operatorname{SiLU}\bigl(
\operatorname{sn}\bigl(\mathbf{W}^{\mathrm{cov}1}_{s(i)}\bigr)\hat{\mathbf{o}}_i + \mathbf{b}^{\mathrm{cov}1}_{s(i)}\bigr)
+ \mathbf{b}^{\mathrm{cov}2}_{s(i)} \in \mathbb{R}^6,
\end{gathered}
\label{eq:methods:dec-heads}
\end{equation}
where $\Delta\bar{\mathbf{x}}_i$ is the mean displacement of protein $i$ and $\boldsymbol{\varpi}_i = (\varpi_{i,1}, \ldots, \varpi_{i,6})$ contains its six raw covariance parameters.
The bias $\mathbf{b}^{\mathrm{dr}2}_s$ is initialized to zero.
The PIEZO system has one set of output heads, and the PIEZO-TREK system has one set per species.
The attention block, the feedforward block and the layer normalizations are shared by all species.

The six covariance parameters of protein $i$ are mapped to a lower-triangular Cholesky factor of the covariance of its displacement,
\begin{equation}
\boldsymbol{\Lambda}_i =
\begin{pmatrix}
\operatorname{softplus}(\varpi_{i,1}) + 10^{-3} & 0 & 0 \\
\varpi_{i,4} & \operatorname{softplus}(\varpi_{i,2}) + 10^{-3} & 0 \\
\varpi_{i,5} & \varpi_{i,6} & \operatorname{softplus}(\varpi_{i,3}) + 10^{-3}
\end{pmatrix},
\label{eq:methods:cholesky}
\end{equation}
in which the softplus floor of $10^{-3}$ keeps every diagonal entry strictly positive and the covariance $\boldsymbol{\Lambda}_i \boldsymbol{\Lambda}_i^{\!\top}$ positive definite.
The three off-diagonal entries are unconstrained and represent the correlations between the $\mathrm{x}$, $\mathrm{y}$ and $\mathrm{z}$ components of the displacement of a single protein.
The covariance is block diagonal across proteins.

Displacements are sampled by reparameterization \cite{kingma2014autoencoding,rezende2014stochastic}, which writes each sample as a deterministic function of the network outputs and of independent standard normal noise,
\begin{equation}
\Delta\mathbf{x}^{(t)}_i = \Delta\bar{\mathbf{x}}_i + \boldsymbol{\Lambda}_i\,
\boldsymbol{\varepsilon}^{(t)}_i,
\qquad
\boldsymbol{\varepsilon}^{(t)}_i \sim \mathcal{N}(\mathbf{0}, \mathbf{I}_3),
\qquad i = 1,\ldots,N,
\label{eq:methods:decoder-sample}
\end{equation}
where $\boldsymbol{\varepsilon}^{(t)}_i$ is a vector of three independent standard normal numbers drawn at every step.
Gradients flow through both the mean and the covariance factor.
The $N$ displacements advance the wrapped state,
\begin{equation}
\mathbf{x}^{(t+1)}_i = \bigl(\mathbf{x}^{(t)}_i + \Delta\mathbf{x}^{(t)}_i\bigr)
\bmod L ,
\label{eq:methods:state-update}
\end{equation}
applied to all three components.
The next geometry vector is computed from $\mathbf{x}^{(t+1)}$ and $\mathbf{x}^{(t)}$.

A training window of $3{,}000$ frames uses frame $0$ to build the first geometry vector, and Equations~\ref{eq:methods:decoder-sample} and \ref{eq:methods:state-update} are then iterated $2{,}999$ times, each iteration starting from the position produced by the previous iteration.

\begin{table}[ht]
\caption{Widths of the internal vectors of the model and parameter counts for each system.}
\label{tab:methods-dims}
\begin{tabular}{@{}lrr@{}}
\toprule
 & PIEZO & PIEZO-TREK \\
\midrule
Geometry vector $\mathbf{e}^{(t)}$          & 462           & 1207 \\
Bank input $\tilde{\mathbf{e}}^{(t)}_i$      & 84            & 199 \\
LSTM banks (one per species)                 & 1             & 2 \\
Bank readout $\boldsymbol{\psi}^{(t)}$       & 3136          & 7616 \\
Transition head hidden width                 & 1568          & 3808 \\
Latent state $\boldsymbol{\zeta}^{(t)}$      & 42            & 102 \\
Decoder tokens (proteins and pair)           & $7 + 1$       & $17 + 1$ \\
Decoder head sets (one per species)          & 1             & 2 \\
Parameters, transformer encoder              & 33{,}075      & 33{,}705 \\
Parameters, LSTM banks and transition head   & 5{,}259{,}954 & 30{,}359{,}842 \\
Parameters, decoder                          & 553{,}265     & 649{,}888 \\
Parameters, latent dynamics                  & 2             & 4 \\
Parameters, total                            & 5{,}846{,}296 & 31{,}043{,}439 \\
\bottomrule
\end{tabular}
\end{table}

\subsection{Training loss}\label{subsec:methods-loss}

The loss compares statistics of the rollout of each training window with the same statistics of frames $1$ to $2{,}999$ of the window, mostly as empirical distributions.
For two samples $\mathbf{y}^{\mathrm{mod}}$ and $\mathbf{y}^{\mathrm{MD}}$ of equal size $K$, the first from the rollout and the second from the MD, the distribution-matching function sorts each sample and computes the mean absolute difference of entries of equal rank,
\begin{equation}
\mathcal{Q}\bigl(\mathbf{y}^{\mathrm{mod}},\, \mathbf{y}^{\mathrm{MD}}\bigr)
= \frac{1}{K}\sum_{n=1}^{K}
\bigl|y^{\mathrm{mod}}_{(n)} - y^{\mathrm{MD}}_{(n)}\bigr| ,
\label{eq:methods:quantile-loss}
\end{equation}
where $y_{(n)}$ is the $n$-th smallest entry of a sample.
For samples of equal size, this quantity is the 1-Wasserstein distance between the two empirical distributions \cite[arXiv version, Remark 2.28]{peyre2019cot}.
The sorting removes the correspondence between individual draws of the two samples.
The rollout and MD samples have the same size, and each rollout has the shape of its target.

Single-protein samples are matched separately within each species.
For a quantity computed at every window, time origin and protein or protein pair, the sample over the rollouts of a batch is denoted $\{\cdot\}^{\mathrm{mod}}$, and the same sample over the corresponding MD target frames is denoted $\{\cdot\}^{\mathrm{MD}}$.
The species-wise match is
\begin{equation}
\mathcal{Q}_{\mathrm{sp}}\bigl(\{\cdot\}^{\mathrm{mod}},\, \{\cdot\}^{\mathrm{MD}}\bigr)
= \frac{1}{|\mathcal{S}|}\sum_{s \in \mathcal{S}}
\mathcal{Q}\bigl(\{\cdot\}^{\mathrm{mod}}_{s},\, \{\cdot\}^{\mathrm{MD}}_{s}\bigr),
\label{eq:methods:quantile-species}
\end{equation}
where the subscript $s$ selects the entries of the proteins of species $s$ and $|\mathcal{S}|$ is the number of species.
Within a species, all windows, time origins and proteins are pooled into one sorted sample.
The match is invariant to relabeling of the proteins within a species and is sensitive to differences in mobility between species.
The species are weighted equally, independent of their numbers of proteins.
For the PIEZO system, which has one species, Equation~\ref{eq:methods:quantile-species} reduces to the pooled match.
Terms based on protein pairs are not split by species and apply Equation~\ref{eq:methods:quantile-loss} to all pairs.

The single-protein lateral observables are computed on the unwrapped lateral positions of Equation~\ref{eq:methods:unwrap}, accumulated from the first predicted frame for the rollout and from frame $1$ for the MD targets.
The lateral displacement over a lag of $\tau$ frames is $\Delta_\tau \mathbf{r}^{(t)}_i = \mathbf{r}^{(t+\tau)}_i - \mathbf{r}^{(t)}_i$.
A statistic at lag $\tau$ uses every time origin $t$ at which it is defined within the $2{,}999$ frames, lags longer than the window are omitted, and an average over a lag set runs over the remaining lags.

Two loss terms constrain the lateral motion of a single protein, the displacement term and the turning-angle term.
These two terms and the cohesion term use the lag set $\mathcal{T}_{\mathrm{lag}} = \{1, 5, 10, 25,\allowbreak 50, 100,\allowbreak 200, 300, 500,\allowbreak 1000, 2000, 3000\}$ in both systems.

The displacement term matches the distribution of displacement magnitudes at each lag,
\begin{equation}
\mathcal{L}_{\mathrm{disp}} = \Bigl\langle
\mathcal{Q}_{\mathrm{sp}}\Bigl(
\Bigl\{\sqrt{\|\Delta_\tau \mathbf{r}^{(t)}_i\|^2 + 10^{-6}}\Bigr\}^{\mathrm{mod}},\;
\Bigl\{\sqrt{\|\Delta_\tau \mathbf{r}^{(t)}_i\|^2 + 10^{-6}}\Bigr\}^{\mathrm{MD}}
\Bigr)\Bigr\rangle_{\tau \in \mathcal{T}_{\mathrm{lag}}},
\label{eq:methods:disp-loss}
\end{equation}
where the constant $10^{-6}$ keeps the square root differentiable at zero displacement.
This term is the distributional counterpart of the time-averaged mean squared displacement.
Lag $3000$ does not fit in the window and is omitted.

The turning-angle term matches the distribution of the unsigned in-plane angle between successive displacements over the same lag \cite{burov2013},
\begin{equation}
\begin{gathered}
\theta^{(t)}_i(\tau) = \operatorname{atan2}\Bigl(
\bigl|\Delta_\tau \mathbf{r}^{(t)}_i \times \Delta_\tau \mathbf{r}^{(t+\tau)}_i\bigr|,\;
\Delta_\tau \mathbf{r}^{(t)}_i \cdot \Delta_\tau \mathbf{r}^{(t+\tau)}_i\Bigr)
\in [0,\pi],
\\
\mathcal{L}_{\mathrm{ang}} = \Bigl\langle
\mathcal{Q}_{\mathrm{sp}}\bigl(\{\theta^{(t)}_i(\tau)\}^{\mathrm{mod}},\,
\{\theta^{(t)}_i(\tau)\}^{\mathrm{MD}}\bigr)
\Bigr\rangle_{\tau \in \mathcal{T}_{\mathrm{lag}}},
\end{gathered}
\label{eq:methods:angle-def}
\end{equation}
where $\mathbf{y}_1 \times \mathbf{y}_2 = y_{1,1}y_{2,2} - y_{1,2}y_{2,1}$ is the scalar cross product of two lateral vectors.
A lag is used when the window yields at least two angles, and lags $2000$ and $3000$ are omitted.

Two loss terms constrain the collective structure of the proteins, the pairwise term and the cohesion term.
Both terms use the lateral minimum-image pair distances of the wrapped coordinates,
\begin{equation}
\ell^{(t)}_{ij} = \sqrt{\bigl\|\operatorname{mi}\bigl(\mathbf{x}^{(t)}_{i,\mathrm{xy}}
- \mathbf{x}^{(t)}_{j,\mathrm{xy}}\bigr)\bigr\|^2 + 10^{-8}}\;\Big/\,L,
\qquad 1 \le i < j \le N,
\label{eq:methods:lateral-pair}
\end{equation}
which differ from the distances of Equation~\ref{eq:methods:pairdist} only by the omission of the membrane-normal component.
The pairwise term matches the distribution of these distances, and the cohesion term matches the distribution of their increments $\Delta_\tau \ell^{(t)}_{ij} = \ell^{(t+\tau)}_{ij} - \ell^{(t)}_{ij}$,
\begin{equation}
\mathcal{L}_{\mathrm{pair}} = \mathcal{Q}\bigl(\{\ell^{(t)}_{ij}\}^{\mathrm{mod}},\,
\{\ell^{(t)}_{ij}\}^{\mathrm{MD}}\bigr),
\qquad
\mathcal{L}_{\mathrm{coh}} = \Bigl\langle\mathcal{Q}\bigl(
\{\Delta_\tau \ell^{(t)}_{ij}\}^{\mathrm{mod}},\,
\{\Delta_\tau \ell^{(t)}_{ij}\}^{\mathrm{MD}}\bigr)\Bigr\rangle_{\tau \in
\mathcal{T}_{\mathrm{lag}}},
\label{eq:methods:pair-loss}
\end{equation}
with samples taken over windows, time origins and all pairs.
Lag $3000$ is omitted from the cohesion term.

The remaining terms use three lateral statistics, which also enter the evaluation of Section~\ref{subsec:methods-stability}.
For a set $\mathcal{P}$ of proteins, these statistics are the mean squared displacement, the non-Gaussian parameter \cite{rahman1964} and the normalized velocity autocorrelation function (VACF),
\begin{align}
\operatorname{MSD}_{\mathcal{P}}(\tau) &= \bigl\langle
\|\Delta_\tau \mathbf{r}^{(t)}_i\|^2 \bigr\rangle,
\label{eq:methods:msd}\\
\alpha_{2,\mathcal{P}}(\tau) &= \frac{\bigl\langle
\|\Delta_\tau \mathbf{r}^{(t)}_i\|^4\bigr\rangle}{
2\bigl\langle\|\Delta_\tau \mathbf{r}^{(t)}_i\|^2\bigr\rangle^2} - 1,
\label{eq:methods:ngp}\\
\operatorname{VACF}_{\mathcal{P}}(\tau) &= \frac{
\bigl\langle \Delta_1 \mathbf{r}^{(t)}_i \cdot \Delta_1 \mathbf{r}^{(t+\tau)}_i \bigr\rangle}{
\bigl\langle \|\Delta_1 \mathbf{r}^{(t)}_i\|^2 \bigr\rangle},
\label{eq:methods:vacf}
\end{align}
where every average runs over the trajectories considered, every time origin $t$ at which the quantity is defined and every protein $i \in \mathcal{P}$, and the constants of at most $10^{-8}$ that guard the denominators are omitted.
The mean squared displacement is time-averaged \cite{metzler2014anomalous}.

The non-Gaussian parameter, denoted $\alpha_2$ by convention and distinct from the exponent $\alpha$ below, uses the factor $2$ for two dimensions.
It vanishes for a Gaussian distribution of displacements, is positive for tails heavier than Gaussian and is negative for suppressed tails.
The VACF uses single-frame displacements as velocities and is normalized by the mean squared single-frame displacement.

Three loss terms constrain the velocity autocorrelation at different levels of detail.
The first term matches the full distribution of the cosine similarity between two displacements over a window of $\tau'$ frames separated by $\tau$ frames,
\begin{equation}
\begin{gathered}
\chi^{(t)}_i(\tau', \tau) = \frac{
\Delta_{\tau'} \mathbf{r}^{(t)}_i \cdot \Delta_{\tau'} \mathbf{r}^{(t+\tau)}_i}{
\sqrt{\|\Delta_{\tau'} \mathbf{r}^{(t)}_i\|^2 + 10^{-8}}\,
\sqrt{\|\Delta_{\tau'} \mathbf{r}^{(t+\tau)}_i\|^2 + 10^{-8}}},
\\
\mathcal{L}_{\mathrm{vq}} = \Bigl\langle \mathcal{Q}_{\mathrm{sp}}\bigl(
\{\chi^{(t)}_i(\tau',\tau)\}^{\mathrm{mod}},\,
\{\chi^{(t)}_i(\tau',\tau)\}^{\mathrm{MD}}\bigr)
\Bigr\rangle_{\tau' \in \mathcal{T}_{\mathrm{vel}},\, \tau \in \mathcal{T}_{\mathrm{corr}}},
\end{gathered}
\label{eq:methods:vacf-def}
\end{equation}
over the velocity windows $\mathcal{T}_{\mathrm{vel}} = \{1, 5, 10, 25, 50, 100, 200\}$ and the correlation lags $\mathcal{T}_{\mathrm{corr}} = \{1, 5, 10, 25,\allowbreak 50, 100,\allowbreak 200, 300, 500,\allowbreak 1000, 2000, 3000\}$.
These two sets are the same for both systems and are independent of $\mathcal{T}_{\mathrm{lag}}$.
Lag $3000$ does not fit in the window, and $77$ of the $84$ pairs of $\tau'$ and $\tau$ are used.

The second VACF term matches the mean normalized autocorrelation curve as a scalar target at each lag,
\begin{equation}
\mathcal{L}_{\mathrm{vc}} = \Bigl\langle \bigl(
\operatorname{VACF}^{\mathrm{mod}}_{\mathrm{all}}(\tau) -
\operatorname{VACF}^{\mathrm{MD}}_{\mathrm{all}}(\tau)\bigr)^2
\Bigr\rangle_{\tau \in \mathcal{T}_{\mathrm{vc}}},
\qquad
\mathcal{T}_{\mathrm{vc}} = \{1, 5, 10, 25, 50, 100, 200\},
\label{eq:methods:vacf-mse}
\end{equation}
where $\operatorname{VACF}_{\mathrm{all}}$ is Equation~\ref{eq:methods:vacf} with $\mathcal{P}$ the set of all proteins, averaged over the windows of the batch and computed on the rollout or on the MD as indicated by the superscript, and $\mathcal{T}_{\mathrm{vc}}$ is its set of correlation lags.
This term does not distinguish species.

The third VACF term compares the raw products $\Pi^{(t)}_i(\tau) = \Delta_1 \mathbf{r}^{(t)}_i \cdot \Delta_1 \mathbf{r}^{(t+\tau)}_i$ point by point, without sorting,
\begin{equation}
\mathcal{L}_{\mathrm{vs}} = \Bigl\langle \bigl\langle
\bigl|\Pi^{(t),\mathrm{mod}}_i(\tau) - \Pi^{(t),\mathrm{MD}}_i(\tau)\bigr|
\bigr\rangle_{\text{windows},\, t,\, i} \Bigr\rangle_{\tau \in \mathcal{T}_{\mathrm{vs}}},
\qquad
\mathcal{T}_{\mathrm{vs}} = \{1, 5, 10, 25, 50, 100\},
\label{eq:methods:vacf-sp}
\end{equation}
where $\Pi^{(t),\mathrm{mod}}_i$ and $\Pi^{(t),\mathrm{MD}}_i$ are the products of the rollout and of the MD at the same window, time and protein.
This term is the only term that compares the rollout and the MD frame by frame.

A rollout and its target share only the initial frame, and the unsorted term is an absolute difference between two independent realizations rather than a distribution match.
For independent realizations, the expectation of the term is smallest when the product of the rollout equals the median of the MD products, and the term favors products with less variability rather than a matching distribution.

Two loss terms target scalar summaries of the diffusive regime.
The first is the anomalous exponent $\alpha$, the slope of the mean squared displacement on logarithmic axes \cite{metzler2000random,metzler2014anomalous}, which distinguishes normal diffusion ($\alpha = 1$), subdiffusion ($\alpha < 1$) and superdiffusion ($\alpha > 1$).
The exponent is obtained by ordinary least squares in closed form, and the gradient of the loss passes through the fit,
\begin{equation}
\begin{gathered}
\alpha_{\mathcal{P}} =
\frac{|\mathcal{T}_{\alpha}|\sum_{\tau} \ln\tau\,\ln\operatorname{MSD}_{\mathcal{P}}(\tau)
- \sum_{\tau}\ln\tau\,\sum_{\tau}\ln\operatorname{MSD}_{\mathcal{P}}(\tau)}{
|\mathcal{T}_{\alpha}|\sum_{\tau}(\ln\tau)^2 - \bigl(\sum_{\tau}\ln\tau\bigr)^2},
\\
\mathcal{T}_{\alpha} = \{1, 5, 10, 25, 50, 100, 200, 300\},
\end{gathered}
\label{eq:methods:alpha-fit}
\end{equation}
with every sum over $\tau \in \mathcal{T}_{\alpha}$ and $|\mathcal{T}_{\alpha}| = 8$ lags.
In the loss, the mean squared displacement is clipped below at $10^{-8}$ before the logarithm and is averaged over a single window and the proteins of one species, and one exponent is fitted per window and species.
The exponent term is
\begin{equation}
\mathcal{L}_{\alpha} = \bigl\langle
\bigl(\alpha^{\mathrm{mod}}_{\mathcal{P}_s} - \alpha^{\mathrm{MD}}_{\mathcal{P}_s}\bigr)^2
\bigr\rangle_{\text{windows},\, s \in \mathcal{S}},
\label{eq:methods:alpha-loss}
\end{equation}
where $\alpha^{\mathrm{mod}}_{\mathcal{P}_s}$ and $\alpha^{\mathrm{MD}}_{\mathcal{P}_s}$ are fitted on the rollout and on the MD target of the same window.

The second scalar term matches the non-Gaussian parameter of Equation~\ref{eq:methods:ngp}.
The non-Gaussian parameter is pooled over the windows of the batch, the time origins and the proteins of a species and is matched by squared error,
\begin{equation}
\mathcal{L}_{\mathrm{ngp}} = \Bigl\langle \frac{1}{|\mathcal{S}|}
\sum_{s \in \mathcal{S}} \bigl(\alpha^{\mathrm{mod}}_{2,\mathcal{P}_s}(\tau)
- \alpha^{\mathrm{MD}}_{2,\mathcal{P}_s}(\tau)\bigr)^2
\Bigr\rangle_{\tau \in \mathcal{T}_{\mathrm{ngp}}},
\qquad
\mathcal{T}_{\mathrm{ngp}} = \{1, 5, 10, 25, 50, 100, 250\}.
\label{eq:methods:ngp-loss}
\end{equation}

The non-Gaussian parameter and the mean squared displacement in the exponent fit have the same definitions in the loss and in the evaluation of Section~\ref{subsec:methods-stability}.
The exponent is used only in the loss.
The loss uses lags that fit within the $2{,}999$ predicted frames of a window, and the evaluation uses lags within rollouts of $40{,}000$ steps.

The total loss is a fixed weighted sum of the nine terms,
\begin{equation}
\begin{split}
\mathcal{L} = {}& w_{\mathrm{disp}}\mathcal{L}_{\mathrm{disp}}
+ w_{\mathrm{ang}}\mathcal{L}_{\mathrm{ang}}
+ w_{\mathrm{pair}}\mathcal{L}_{\mathrm{pair}}
+ w_{\mathrm{coh}}\mathcal{L}_{\mathrm{coh}}
+ w_{\mathrm{vq}}\mathcal{L}_{\mathrm{vq}}\\
&+ w_{\mathrm{vc}}\mathcal{L}_{\mathrm{vc}}
+ w_{\mathrm{vs}}\mathcal{L}_{\mathrm{vs}}
+ w_{\alpha}\mathcal{L}_{\alpha}
+ w_{\mathrm{ngp}}\mathcal{L}_{\mathrm{ngp}},
\end{split}
\label{eq:methods:total-loss}
\end{equation}
with the loss weights $w$ of Table~\ref{tab:methods-weights}.
All terms are lateral, and no term constrains the membrane-normal coordinate.

\begin{table}[ht]
\caption{Loss weights of Equation~\ref{eq:methods:total-loss}. The weights are dimensionless multipliers on observables in normalized units and are identical for the five training seeds of a system.}
\label{tab:methods-weights}
\begin{tabular}{@{}llrr@{}}
\toprule
Term & Weight & PIEZO & PIEZO-TREK \\
\midrule
Displacement           & $w_{\mathrm{disp}}$ & 1.836   & 7.712 \\
Turning angle          & $w_{\mathrm{ang}}$  & 181.67  & 858.79 \\
Pairwise distance      & $w_{\mathrm{pair}}$ & 18.125  & 18.125 \\
Cohesion               & $w_{\mathrm{coh}}$  & 1871.69 & 583.36 \\
VACF distribution      & $w_{\mathrm{vq}}$   & 12.705  & 2.229 \\
VACF curve             & $w_{\mathrm{vc}}$   & 166.43  & 166.43 \\
VACF, unsorted         & $w_{\mathrm{vs}}$   & 4.020   & 4.020 \\
Anomalous exponent     & $w_{\alpha}$        & 2214.83 & 1571.96 \\
Non-Gaussian parameter & $w_{\mathrm{ngp}}$  & 94.555  & 179.43 \\
\bottomrule
\end{tabular}
\end{table}

The weights of the PIEZO system were selected by Gaussian-process Bayesian optimization \cite{snoek2012practical} over $30$ trials in Optuna \cite{akiba2019optuna}.
The first $12$ trials followed a Sobol design \cite{sobol1967distribution}, a low-discrepancy sequence that distributes points evenly over the search space.
Each trial trained one model on the PIEZO system and scored it with the replica-referenced objective of the training monitor (Section~\ref{subsec:methods-schedule}).
The trial with the best score was retained.
The search range of each weight was set separately for each term.

The weight search used a configuration of the model and of the training that differs from the final configuration.
In the search configuration, the decoder used the latent state as the attention query, the transformer encoder received the sine-cosine embedding of absolute position at every frame, and the latent parameters were initialized to $a_s = 0.5$ and $\kappa_s = 0.3$.
The search training used a plateau learning-rate schedule, which halves the learning rate when the monitored objective stops improving, and early stopping, which ends training after the objective has not improved for a set number of epochs.
It also used a learning rate searched between $10^{-7}$ and $10^{-4}$, two rollouts per window and windows of $3{,}000$ frames cut at any position in augmented views of the first $9{,}999$ frames of replica 1, taken before the removal of duplicated frames.
The search loss also included two membrane-normal terms and a latent-autocorrelation term, which the final models do not use.
The nine weights of Table~\ref{tab:methods-weights} are carried over unchanged from the selected trial.

The weights of the PIEZO-TREK system are a point of a $24$-point Latin-hypercube design \cite{mckay1979comparison} around the PIEZO weights.
The design multiplies six weights, those of the displacement, turning-angle, cohesion, VACF-distribution, exponent and non-Gaussian terms, by factors between $0.1$ and $10$ distributed uniformly in their logarithm, and keeps the pairwise, VACF-curve and unsorted VACF weights.
The factors of the selected point are $4.20$, $4.73$, $0.31$, $0.18$, $0.71$ and $1.90$, in the order of the six terms listed above.

Five points of the design, points $1$, $6$, $8$, $11$ and $19$, were trained for the PIEZO-TREK system with training seed $500$.
Every epoch was scored with the rollouts of Section~\ref{subsec:methods-stability} against per-lag replica tolerances taken from the PIEZO system.
Point $11$, which has the lowest best-epoch score under these tolerances, is used.
Seed $500$ is also one of the five training seeds of the final models.

\subsection{Optimization and schedule}\label{subsec:methods-schedule}

The parameters are optimized with Sophia \cite{liu2023sophia}, a stochastic second-order optimizer that divides a moving average of the gradient by a moving average of a diagonal estimate of the Hessian and clips the step of each coordinate.
The two published variants of Sophia estimate this diagonal with the Hutchinson estimator \cite{hutchinson1989trace} or with the Gauss-Newton-Bartlett estimator, which samples labels from the predictive distribution of the model.

We use the clipped update of Sophia with a different curvature estimate, a moving average of the squared minibatch gradient that is refreshed at every step, without sampled labels and without a refresh interval.
For each scalar parameter $\varphi$, with $\partial\mathcal{L}/\partial\varphi$ its minibatch gradient after clipping, each step updates two moving averages $\mathcal{R}_\varphi$ and $\mathcal{G}_\varphi$, both initialized at zero, and then the parameter,
\begin{align}
\mathcal{R}_\varphi &\leftarrow 0.99\,\mathcal{R}_\varphi +
0.01\,\bigl(\partial\mathcal{L}/\partial\varphi\bigr)^2,
\label{eq:methods:sophia-h}\\
\mathcal{G}_\varphi &\leftarrow 0.965\,\mathcal{G}_\varphi +
0.035\,\partial\mathcal{L}/\partial\varphi,
\label{eq:methods:sophia-m}\\
\varphi &\leftarrow \varphi - \eta\,\operatorname{sign}(\mathcal{G}_\varphi)
\min\!\left(\frac{|\mathcal{G}_\varphi|}{\max\bigl(1.017\,\mathcal{R}_\varphi,\,
10^{-15}\bigr)},\; 1\right),
\label{eq:methods:sophia-step}
\end{align}
where $\eta$ is the learning rate of the parameter group and $1.017$ is the product of the curvature parameter $0.015890$ and the batch size $64$.
The step of every coordinate is at most $\eta$, and this bound is reached when the momentum is large compared with the curvature estimate.

The parameters are divided into three groups.
The first group receives weight decay and contains only the weights of the plain linear and attention layers, with $0.58$ million parameters in the PIEZO model and $0.93$ million parameters in the PIEZO-TREK model.
Parameters of this group are multiplied by $1 - 0.01\,\eta$ before the update of Equation~\ref{eq:methods:sophia-step}, which is a decoupled weight decay \cite{loshchilov2019decoupled} of $0.01$.
The second group receives no weight decay and contains every other parameter outside the third group, including the recurrent weights, the weights of the spectrally normalized layers, and all biases and normalization parameters.
The third group contains the latent dynamics parameters $\tilde{a}_s$ and $\tilde{\kappa}_s$ of Equation~\ref{eq:methods:dynamics-params}, receives no weight decay and has a learning rate multiplied by $3162.27$.

The learning rate is constant during training, with $\eta = 2.2117\times10^{-5}$ for the network and $0.06994$ per step for the latent dynamics group.
The base rate was selected on the PIEZO system from the grid $6.994\times10^{-6}$, $2.2117\times10^{-5}$, $6.994\times10^{-5}$ and $2.2117\times10^{-4}$, with the learning rate of the latent dynamics group held at $0.03497$ per step.
The final models use twice this latent learning rate.

Every model is trained for $25$ epochs of $15$ steps, $375$ optimizer steps in total, on full windows of $3{,}000$ frames from the first step.
Five models are trained per system with the training seeds $500$ to $504$.
The seed sets the initial weights, the order of the training windows and the noise of the training rollouts, up to non-deterministic GPU kernels.
Table~\ref{tab:methods-training} lists these settings.

\begin{table}[ht]
\caption{Training settings, shared by both systems.}
\label{tab:methods-training}
\begin{tabular}{@{}ll@{}}
\toprule
Setting & Value \\
\midrule
Optimizer                      & Sophia with squared-gradient curvature, \\
                               & Equations~\ref{eq:methods:sophia-h} to \ref{eq:methods:sophia-step} \\
Learning rate $\eta$           & $2.2117\times10^{-5}$, constant \\
Learning rate, latent dynamics & $0.06994$, constant \\
Weight decay                   & $0.01$, plain linear and attention weights only \\
Batch                          & 64 windows \\
Window                         & 3{,}000 frames, 1 initial condition and 2{,}999 targets \\
Windows per epoch              & 960, giving 15 optimizer steps \\
Epochs                         & 25, giving 375 optimizer steps \\
Gradient clipping              & global norm 20 \\
Dropout                        & 0.252, decoder attention and feedforward only \\
Precision                      & bfloat16 mixed precision, latent update in single \\
                               & precision and rounded to bfloat16 between steps \\
Training seeds                 & 500 to 504 \\
\bottomrule
\end{tabular}
\end{table}

Gradients are computed through the whole $2{,}999$-step rollout of a window by backpropagation through time \cite{werbos1990backpropagation}, which takes the gradient of a recurrent computation through all of its steps.
Training uses bfloat16 mixed precision and allows reduced-precision arithmetic in single-precision matrix products.
Gradients are clipped to a global norm of $20$ \cite{pascanu2013difficulty}.

Models were trained with PyTorch 2.10.0 \cite{paszke2019pytorch} and PyTorch Lightning 2.6.1 \cite{falcon2019lightning} on NVIDIA L40, L40S and RTX A6000 GPUs.

At the end of every epoch, a training monitor rolls out $128$ initial conditions from the reference replicas for $20{,}000$ steps and computes a summary objective.
The objective is the mean, over six lateral statistics and, in the PIEZO-TREK system, a seventh statistic that compares the mobilities of the two species, of the rectified excess of the distance between the model and the MD relative to the distance between two MD replicas.
For the PIEZO-TREK system, the replica distance is taken from the PIEZO system.
The monitor changes neither the gradients nor the learning rate, does not enter the model selection of Section~\ref{subsec:methods-selection}, and enters training only through the stopping rule above.
The weight search of Section~\ref{subsec:methods-loss} was scored with an objective of the same form.

\subsection{Long-horizon evaluation}\label{subsec:methods-stability}

A single long-horizon evaluation is used for model selection and for reporting.
The evaluation measures the agreement between the lateral diffusion statistics of long rollouts and those of the reference trajectories.
It is applied, separately from training, to the weights at the end of every epoch of every training run.
Each rollout starts from one MD configuration and is generated as described in Section~\ref{subsec:methods-overview}.

For each system, $512$ initial conditions are drawn without replacement, with a generator of fixed seed, from the pooled frames of the replicas listed in Table~\ref{tab:methods-eval}.
Each drawn frame supplies the positions of all proteins for one initial condition.
The same $512$ initial conditions are used for every checkpoint of a system, and most of them lie outside the training frames.

Each initial condition is rolled out autoregressively for $40{,}000$ steps under Equations~\ref{eq:methods:decoder-sample} and \ref{eq:methods:state-update}, with dropout disabled and in single precision.
For the PIEZO system, $40{,}000$ steps correspond to $20$~\textmu s.
The rollout length is $3.2$ times the frame count of the longest PIEZO replica and $3.6$ times that of the longest PIEZO-TREK replica.
Both the latent integrator and the decoder are stochastic, and each checkpoint is rolled out under three noise seeds with the initial conditions held fixed, which gives $512 \times 40{,}000 \times 3 = 6.14\times10^{7}$ predicted frames per checkpoint.

\begin{table}[ht]
\caption{Evaluation settings. Initial conditions are drawn from the listed frames, and the reference pools both replicas of each system. The largest scored lag $\tau_{\max,s}$ of species $s$ is defined in the text, and the stand-in rollouts and the draws are defined in Section~\ref{subsec:methods-allowance}.}
\label{tab:methods-eval}
\begin{tabular}{@{}lll@{}}
\toprule
 & PIEZO & PIEZO-TREK \\
\midrule
Initial conditions from     & replicas 1 and 2 & replica 1 \\
Frames available            & 19{,}681         & 11{,}001 \\
Largest lag $\tau_{\max,s}$ & 5000             & 5000 for both species \\
Cells                       & 204              & $204 + 204 = 408$ \\
Stand-in rollouts           & $2\times40$      & $2\times40$ \\
Draws                       & 600              & 960 \\
\bottomrule
\end{tabular}
\end{table}

The statistics are computed separately for each species $s$ on the lateral coordinates at the lags of the ladder $\mathcal{T}_s = \{1, 100, 200, \ldots, \tau_{\max,s}\}$, which contains lag $1$ and every multiple of $100$ up to $\tau_{\max,s}$.
Lags $2$ to $99$ are not scored.
At each lag, the evaluation computes the mean squared displacement, the non-Gaussian parameter and the VACF of Equations~\ref{eq:methods:msd} to \ref{eq:methods:vacf} with $\mathcal{P} = \mathcal{P}_s$, and the empirical sample of displacement magnitudes
\begin{equation}
\mathcal{V}_s(\tau) = \bigl\{\, \|\Delta_\tau \mathbf{r}^{(t)}_i\| \;:\;
\text{all trajectories, } t,\; i \in \mathcal{P}_s \,\bigr\},
\label{eq:methods:vanhove}
\end{equation}
which is the empirical form of the self part of the van Hove correlation function \cite{vanhove1954}.
For the rollouts, all $512$ rollouts, all time origins and all proteins of the species are pooled after unwrapping with Equation~\ref{eq:methods:unwrap}.

A sample with more than $120{,}000$ values is reduced to $120{,}000$ values by a uniform draw with replacement, with a fixed seed for each lag.

The reference statistics are computed from the MD replicas in the same way, replica by replica.
The reference value of the mean squared displacement, the non-Gaussian parameter and the VACF is the mean of the two replica values, and the reference van Hove sample is the concatenation of the two replica samples.
Replica 1 of each system contains the $3{,}000$ training frames, and the reference includes these frames.

Each statistic $M \in \{\mathrm{MSD}, \mathrm{NGP}, \mathrm{vH}, \mathrm{VACF}\}$ of species $s$ at lag $\tau$ forms one cell, and the distance between the model and the reference in that cell is
\begin{equation}
\delta_{s,M}(\tau) =
\begin{cases}
\Bigl|\ln\dfrac{\max\bigl(\operatorname{MSD}^{\mathrm{mod}}_{\mathcal{P}_s}(\tau),\,10^{-9}\bigr)}{
\max\bigl(\operatorname{MSD}^{\mathrm{ref}}_{\mathcal{P}_s}(\tau),\,10^{-9}\bigr)}\Bigr|,
& M = \mathrm{MSD},\\[12pt]
\bigl|\alpha^{\mathrm{mod}}_{2,\mathcal{P}_s}(\tau) - \alpha^{\mathrm{ref}}_{2,\mathcal{P}_s}(\tau)\bigr|,
& M = \mathrm{NGP},\\[6pt]
\bigl|\operatorname{VACF}^{\mathrm{mod}}_{\mathcal{P}_s}(\tau) - \operatorname{VACF}^{\mathrm{ref}}_{\mathcal{P}_s}(\tau)\bigr|,
& M = \mathrm{VACF},\\[6pt]
W_1\bigl(\mathcal{V}^{\mathrm{mod}}_s(\tau),\, \mathcal{V}^{\mathrm{ref}}_s(\tau)\bigr)
\big/ \bigl(\operatorname{rms}\bigl(\mathcal{V}^{\mathrm{ref}}_s(\tau)\bigr) + 10^{-6}\bigr),
& M = \mathrm{vH},
\end{cases}
\label{eq:methods:celldist}
\end{equation}
where the superscripts $\mathrm{mod}$ and $\mathrm{ref}$ denote the rollout statistic and the reference statistic, and $\operatorname{rms}(\cdot)$ is the root mean square of a sample.
The 1-Wasserstein distance $W_1$ between two samples $\mathcal{Y}$ and $\mathcal{Y}'$ is evaluated at evenly spaced probability levels,
\begin{equation}
W_1(\mathcal{Y}, \mathcal{Y}') = \frac{1}{J}\sum_{n=0}^{J-1}
\Bigl|F^{-1}_{\mathcal{Y}}\Bigl(\frac{n}{J-1}\Bigr) -
F^{-1}_{\mathcal{Y}'}\Bigl(\frac{n}{J-1}\Bigr)\Bigr|,
\qquad J = \min\bigl(|\mathcal{Y}|, |\mathcal{Y}'|, 120{,}000\bigr),
\label{eq:methods:w1}
\end{equation}
with $F^{-1}_{\mathcal{Y}}$ the empirical quantile function of $\mathcal{Y}$, linearly interpolated between order statistics, and $|\mathcal{Y}|$ the size of a sample.
The two samples differ in size, and the probability levels replace the rank pairing of Equation~\ref{eq:methods:quantile-loss}.

For the PIEZO-TREK system, the per-epoch evaluation computes the distances of the model to replica 1 alone, and the score uses the reference pooled over both replicas.
For the mean squared displacement, the non-Gaussian parameter and the VACF, the model value of each rollout seed is recovered from its distance to replica 1, the value of replica 1 and the mean model curve, and the distance to the pooled reference is computed from this value.
The per-epoch evaluation of this system provides the van Hove distance of the model to replica 1 but not the van Hove sample of the model.
The van Hove distance to the pooled reference is replaced by an upper bound from the triangle inequality of $W_1$,
\begin{equation}
\delta_{s,\mathrm{vH}}(\tau) \le
\frac{W_1\bigl(\mathcal{V}^{\mathrm{mod}}_s, \mathcal{V}^{\mathrm{rep1}}_s\bigr) +
W_1\bigl(\mathcal{V}^{\mathrm{rep1}}_s, \mathcal{V}^{\mathrm{ref}}_s\bigr)}{
\operatorname{rms}\bigl(\mathcal{V}^{\mathrm{ref}}_s\bigr) + 10^{-6}},
\label{eq:methods:trek-bound}
\end{equation}
where $\mathcal{V}^{\mathrm{rep1}}_s$ is the van Hove sample of replica 1 alone and the lag argument $\tau$ is omitted on the right.
The score of a PIEZO-TREK checkpoint uses a value at least as large as the true van Hove distance, up to the quantile discretization of Equation~\ref{eq:methods:w1}.

\subsection{Perfect-model allowance and excess score}\label{subsec:methods-allowance}

The allowance of a cell is the distance that a perfect model reaches or exceeds with a probability of $5$ percent, given a reference of the same size and layout as the MD reference.
A reference built from two finite replicas differs from the long-run statistics of the process that produced it, and a perfect model has a nonzero distance in every cell.

Two trained models of each system stand in for the true process.
These models are distinct from the evaluated models and come from separate training runs with other settings.
Each stand-in model is rolled out from its own initial conditions for $40{,}000$ steps, with $40$ rollouts per model, and its rollouts are divided into two halves.
The long-run statistics of one half, computed from all rollouts at full length, serve as the true statistics.
The construction is a parametric bootstrap \cite{efron1993bootstrap}, in which replicate references are simulated from fitted models instead of being resampled from the MD.

Each draw builds a reference from the other half in the layout of the MD reference.
For each MD replica, the draw takes a distinct rollout and cuts from it a contiguous segment of the length of that replica at a random offset.
The cut reference is assembled in the same way as the MD reference, and its distance to the true statistics, $\delta^{\mathrm{draw}}_{s,M}(\tau)$, computed with Equation~\ref{eq:methods:celldist}, measures the deviation introduced by a finite reference alone.
Both halves serve in turn as the true statistics, and the draws of the two stand-in models are pooled, which gives $600$ draws for the PIEZO system and $960$ draws for the PIEZO-TREK system.

The raw allowance $\tilde{A}_{s,M}(\tau)$ is the 95th percentile of $\delta^{\mathrm{draw}}_{s,M}(\tau)$ over the draws.
For lags of $100$ and above, the raw allowance is smoothed along the ladder.
The smoothed allowance at a lag is the value at that lag of a straight line fitted by least squares to $\ln\tilde{A}$ against $\ln\tau$ over the five ladder points centered on it, with lag $1$ excluded and with fewer points at the ends of the ladder, but never fewer than three.

At lag $1$, the allowance is the larger of $\tilde{A}$ and the 95th percentile of the same distance between the MD reference and $400$ of its circular block-bootstrap resamples \cite{kunsch1989jackknife,politis1992circular}.
The bootstrap resamples each MD replica in blocks of $1{,}000$ frames with uniformly placed starts, wraps around the end of the replica and moves whole frames, which keeps all proteins of a frame together.
The result is the allowance $A_{s,M}(\tau)$.

In cells where the distance between the two MD replicas exceeds both the 95th percentile of the corresponding distance between draws and the allowance, the allowance is replaced by the distance between the replicas.
With $\Gamma_{s,M}(\tau)$ the distance of Equation~\ref{eq:methods:celldist} between the two MD replicas and $\Gamma^{\mathrm{draw}}_{s,M}(\tau)$ the same distance between the two replica cuts of one draw, the guarded allowance is
\begin{equation}
A^{\mathrm{g}}_{s,M}(\tau) =
\begin{cases}
\Gamma_{s,M}(\tau) & \text{if } \Gamma_{s,M}(\tau) > \operatorname{pct}_{95}\bigl(\Gamma^{\mathrm{draw}}_{s,M}(\tau)\bigr)
\text{ and } \Gamma_{s,M}(\tau) > A_{s,M}(\tau),\\[4pt]
A_{s,M}(\tau) & \text{otherwise,}
\end{cases}
\label{eq:methods:guard}
\end{equation}
where $\operatorname{pct}_{95}$ is the 95th percentile over the draws.

For one checkpoint and one rollout seed, the excess score is the mean, over all cells of the system, of the rectified excess of the distance over its allowance,
\begin{equation}
\mathcal{D} = \frac{1}{|\mathcal{C}|}\sum_{(s,M,\tau) \in \mathcal{C}}
\max\!\left(0,\;
\frac{\delta_{s,M}(\tau)}{\max\bigl(A^{\mathrm{g}}_{s,M}(\tau),\,10^{-4}\bigr)} - 1\right),
\label{eq:methods:excess}
\end{equation}
where $\mathcal{C}$ is the set of cells, consisting of the four statistics at every lag of $\mathcal{T}_s$ for every species $s$, and $|\mathcal{C}|$ is its size, $204$ for the PIEZO system and $408$ for the PIEZO-TREK system.
All cells have the same weight.
The score of a checkpoint, $\bar{\mathcal{D}}$, is the mean of $\mathcal{D}$ over its three rollout seeds.
A lower score indicates closer agreement with the reference, and $\bar{\mathcal{D}} = 0$ when every cell lies within its allowance.
The score of a trained model uses the guarded allowance $A^{\mathrm{g}}_{s,M}(\tau)$, which is built from the draws of both stand-in models and widened at lag $1$ by the bootstrap.

The threshold $\mathcal{D}_{95}$ of a system is the 95th percentile of the excess score that the stand-in models obtain out of sample.
Each draw of a stand-in model is scored once with Equation~\ref{eq:methods:excess}, against its own cut reference and against allowances and a guard built from the draws of the other stand-in model alone, without the bootstrap widening at lag $1$.
No selection over epochs or seeds enters the threshold.
A checkpoint is at or below the threshold when $\bar{\mathcal{D}} \le \mathcal{D}_{95}$.

\subsection{Model selection}\label{subsec:methods-selection}

The model of each training seed is the checkpoint of its best epoch.
For each of the five training seeds of a system, the checkpoints of all $25$ epochs are evaluated with the protocol of Sections~\ref{subsec:methods-stability} and \ref{subsec:methods-allowance}, and the selected epoch is
\begin{equation}
\hat{E} = \operatorname*{arg\,min}_{E \in \{0, \ldots, 24\}}
\bar{\mathcal{D}}_E,
\label{eq:methods:selection}
\end{equation}
where $E$ is the zero-based epoch index, the checkpoint of epoch $E$ has been trained for $15(E+1)$ optimizer steps, and $\bar{\mathcal{D}}_E$ is its score.
This procedure gives five models per system and ten models in total, each at its own best epoch.

The score used to select an epoch and the score reported for the selected model come from the same evaluation.
Both use the same $512$ initial conditions, the same three rollout seeds and the same reference, which pools both replicas of the system and includes the training frames.
No validation loss is used, and the training loss does not enter the selection.

The selection is also repeated with one rollout seed held out.
The epoch of each run is chosen from the mean of $\mathcal{D}$ over two of its three rollout seeds and scored on the remaining seed, for each of the three choices of the held-out seed.
The initial conditions and the reference are the same for all three seeds, and this check holds out rollout noise and not data.

The threshold $\mathcal{D}_{95}$ depends on the reference layout and on the stand-in models and does not depend on the trained model.

\subsection{Metrics}\label{subsec:methods-metrics}

The time-averaged mean squared displacement (TAMSD) of a protein is its squared lateral displacement averaged over all time origins at a given lag time $\tau$.
For a single trajectory of $T$ frames with unwrapped lateral positions $\mathbf{r}(t)$, the TAMSD is
\begin{equation}\label{eq:TAMSD}
\overline{\Delta r^2}(\tau) =
\frac{1}{T-\tau}
\sum_{t=0}^{T-\tau-1}
\bigl\| \mathbf{r}(t+\tau) - \mathbf{r}(t) \bigr\|^2,
\end{equation}
where the squared displacement is $\bigl(x(t+\tau)-x(t)\bigr)^2 + \bigl(y(t+\tau)-y(t)\bigr)^2$, with $x$ and $y$ the lateral coordinates.
The TAMSD is the mean squared displacement of Equation~\ref{eq:methods:msd} for a single protein and a single trajectory.

For normal diffusion in two dimensions, the ensemble average of the TAMSD grows linearly with the lag time, $\langle\overline{\Delta r^2}(\tau)\rangle = 4D\tau$, and the diffusion coefficient $D$ is obtained from this relation.
The anomalous exponent $\alpha$ is obtained by a fit of the TAMSD curves to
\begin{equation}\label{eq:alpha}
 \overline{\Delta r^2}(\tau) \propto \tau^{\alpha}
\end{equation}

The motion is normal diffusion for $\alpha = 1$, subdiffusion for $\alpha < 1$ and superdiffusion for $\alpha > 1$.

The self part of the van Hove function, $G_s(r,\tau)$, is the probability density of the lateral displacement of a protein over a lag time $\tau$, expressed as a function of the displacement magnitude $r$,
\begin{equation}\label{eq:vanhove}
G_s(r,\tau) =
\frac{1}{2\pi r}
\left\langle
\delta\!\left(r - \sqrt{\Delta x^2 + \Delta y^2}\right)
\right\rangle,
\end{equation}
where $\Delta x = x(t_0+\tau)-x(t_0)$ and $\Delta y = y(t_0+\tau)-y(t_0)$, and the brackets denote an average over all time origins $t_0$ and all trajectories of the ensemble.
The function satisfies the radial normalization $\int_0^{\infty} 2\pi r\, G_s(r,\tau)\,dr = 1$.
Equation~\ref{eq:methods:vanhove} gives the empirical sample from which this function is estimated.
The results compare the van Hove functions of the MD and of the model at several lag times.

For Gaussian displacements, the self part of the van Hove function is
\begin{equation}\label{eq:gaussfit}
G_s(r,\tau) = \frac{1}{2\pi \sigma^2(\tau)} \exp\left(-\frac{r^2}{2\sigma^2(\tau)}\right),
\end{equation}
where $\sigma^2(\tau)$ is the variance of each Cartesian component of the lateral displacement at lag time $\tau$.

The non-Gaussian parameter measures the deviation of the distribution of lateral displacements from a Gaussian distribution,
{\setlength{\abovedisplayskip}{4pt}
\setlength{\belowdisplayskip}{4pt}
\begin{equation}\label{eq:NGP}
\alpha_2(\tau) =
\frac{\langle r^4(\tau) \rangle}
{2\langle r^2(\tau) \rangle^2} - 1,
\end{equation}}
where $r^2(\tau) = \Delta x^2 + \Delta y^2$, $r^4(\tau) = (\Delta x^2 + \Delta y^2)^2$, and the averages run over all proteins and time origins at lag time $\tau$.
This quantity is the non-Gaussian parameter of Equation~\ref{eq:methods:ngp}.
For Gaussian displacements in two dimensions, $\langle r^4\rangle = 2\langle r^2\rangle^2$ and $\alpha_2 = 0$.
Positive values correspond to tails heavier than Gaussian, and negative values correspond to tails narrower than Gaussian.

\section{Results}
\subsection{7 PIEZO}
In a previous publication of our lab ~\cite{thai2026}, we reported VAR  models succeed in describing short time dynamics of a multi-protein (7 PIEZO) system diffusion in the membrane but fails to describe mid/long term dynamics. As mentioned in the methods section for this particular multi-protein system, we improved the forecasting in trajectories, by introducing our new moel architecture described in Methods.

\begin{figure}
    \centering
    \includegraphics[width=1.0\linewidth]{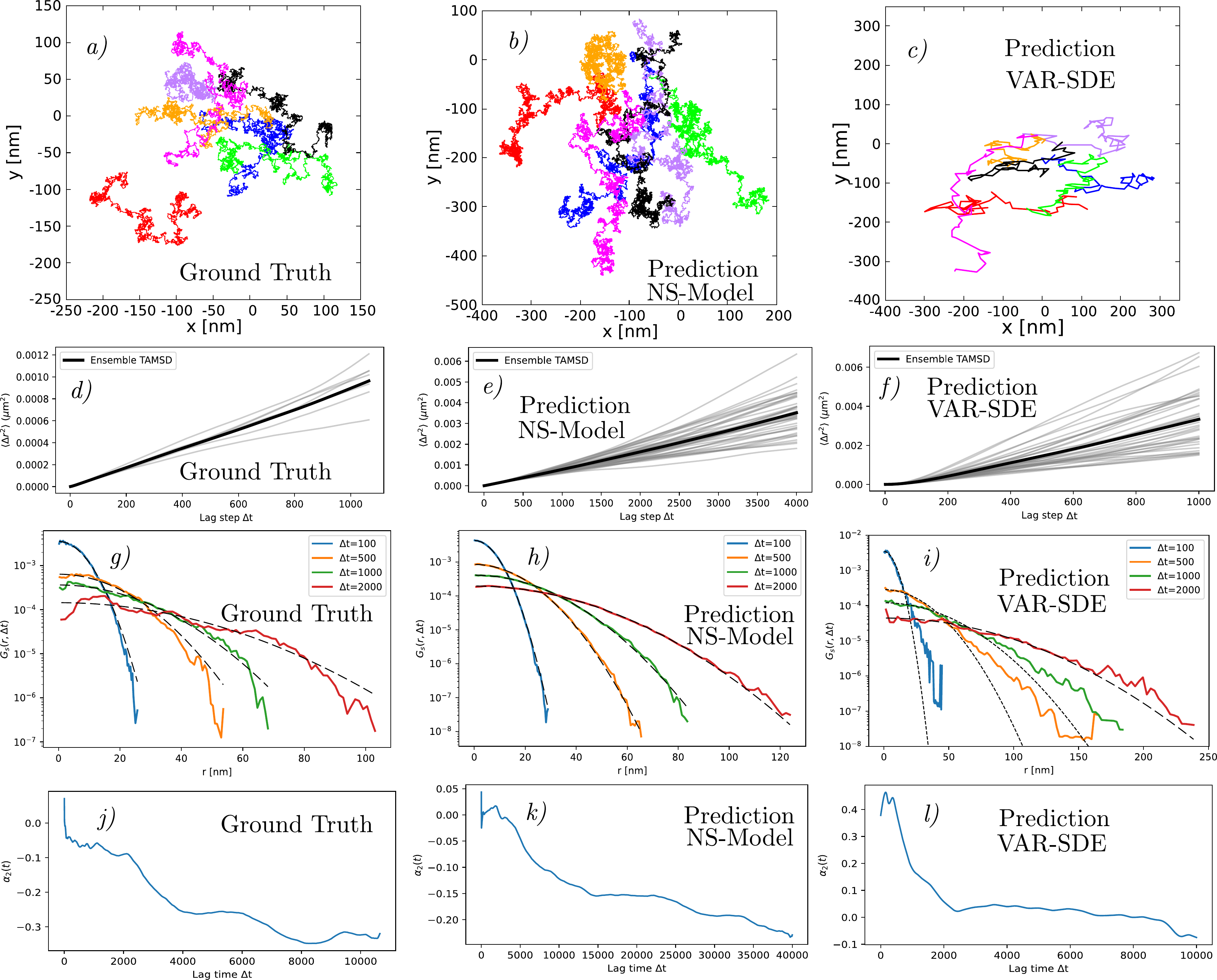}
    \caption{Comparison between the ground truth (MD simulations) and predictions of our neural stochastic (NS) model and VAR-SDE model \cite{thai2026}, for the 7 PIEZO multi-protein system. \textit{a)} Trajectories of the 7 PIEZO channels (center of mass) obtained from MD simulations during a total time of 10,651 time frames. \textit{b)} Predicted trajectories using the NS model of the PIEZO channels during a total time with 40,000 time frames. \textit{c)} Predicted trajectories obtained by  the VAR-SDE model for tracks with 10,000 time frames. \textit{d-f)} TAMSD and ETAMSD for ground truth trajectories and predictions respectively. For convergence of the statistics, the TAMSD is computed in 10\% of the trajectories (cite). \textit{g-i)} Van Hove function Eq.~\ref{eq:vanhove} for selected lag-times $\Delta t =\lbrace 100,500,1000,2000\rbrace$, for MD simulations and predictions. And \textit{j-l)} Non-Gaussian parameter Eq.~\ref{eq:NGP} for lag times covering the whole trajectories of the ground truth and each prediction respectively.     }
    \label{fig:7Piezo}
\end{figure}

Now, as we can see in Fig.~\ref{fig:7Piezo}
we compare the ground truth trajectories
and diffusive statistics with those obtained from our predictions using our physics informed model and the best model from the VAR collection published in ~\cite{thai2026}, namely the VAR-
SDE model with a per-step latent resampling with non-recurrent decoder. In Fig.~\ref{fig:7Piezo}~\textit{a)} the whole unwrapped trajectories are displayed, for the MD simulations (ground truth with 10,651 time frames). The same is shown for the NS model prediction  Fig.~\ref{fig:7Piezo}~\textit{b)} and for those of the VAR-SDE model Fig.~\ref{fig:7Piezo}~\textit{c)}. In the case of the NS predictions the trajectories were sampled  during an extent of 4 more times   compared with the ground truth, \textit{e.g.} 40,000 time frames. For the case of the VAR-SDE model the trajectories are just shown for 10,000 time frames, since they only attempted to replicate the dynamics of the Ground Truth. As we observe the NS model tracks visually resemble more the Ground Truth's, Brownian like motion, when compared with those obtained by using the VAR-SDE model.

In Fig.~\ref{fig:7Piezo}~\textit{d-f)} we show the time-averaged mean-square displacement (TAMSD) Eq.~\ref{eq:TAMSD} and the \textit{ensemble} TAMSD (ETAMSD) for the respective set of trajectories (Ground Truth, NS Model, VAR-SDE model). Having that for the Ground Truth there are 7 trajectories, for the NS Model 42 tracks, and for the VAR-SDE model 35 trajectories. The mean anomalous exponent $\alpha$ defined by Eq.~\eqref{eq:alpha} was obtained by fitting  each TAMSD curve and computing the mean value for each case. For the Ground Truth the mean anomalous exponent is $\bar{\alpha}=1.049 \pm 0.022$ (SEM), for the NS Model we have $\bar{\alpha}=1.048 \pm 0.012$ (SEM), and for the VAR-SDE model we reported $\bar{\alpha}=1.563 \pm 0.016$ (SEM). So we can conclude that the NS model accurately describes the diffusion regime in the seven PIEZO system, and predicts the existence of normal diffusion over 4 times the actual Ground Truth's trajectories.

 The Van Hove functions Eq.\ref{eq:vanhove} (see Fig.~\ref{fig:7Piezo}~\textit{g-i}) show that the NS model better reproduces the displacement distributions over different lag times. Namely compared with the VAR prediction for which the Van Hove distribution exhibits large tails, the corresponding function for the NS model for the lag times analyzed $\Delta t = \lbrace 100,500,1000,2000\rbrace$ remain Gaussian (see Eq.~\ref{eq:gaussfit}), curbing the prescence of considerably large displacements.  And further the  non-Gaussian parameter Eq.~\ref{eq:NGP} (see Fig.~\ref{fig:7Piezo}~\textit{j-l}) provides additional evidence that the  NS model  is able  to capture  the underlying dynamics. In particular the non Gaussian Parameter in the NS prediction Fig.~\ref{fig:7Piezo}~\textit{k)} for lag times $\Delta t>4000$ takes negative values implying the presence of suppressed tails in accordance to the respective metric  in the MD simulation see  Fig.~\ref{fig:7Piezo}~\textit{j)}. Overall, the results in Fig.~\ref{fig:7Piezo} demonstrate that our NS model provides a substantially improved description of the multi-protein diffusion dynamics compared with the previously developed VAR-SDE architecture. In particular, the close agreement between the TAMSD curves of the ground truth and NS model is reflected in their anomalous diffusion exponents, being in all cases $\alpha \sim 1$, contrasting with the super-diffusive behavior rendered by the VAR-SDE model.
 Taken together, these results indicate that our model successfully captures multiple aspects of the transport statistics of the seven PIEZO system, extending the predictive time horizon to approximately four times the duration of the ground-truth trajectory while maintaining the observed diffusive behavior.

\subsection{7 PIEZO and 10 TREK}

\begin{figure}
    \centering
    \includegraphics[width=0.8\linewidth]{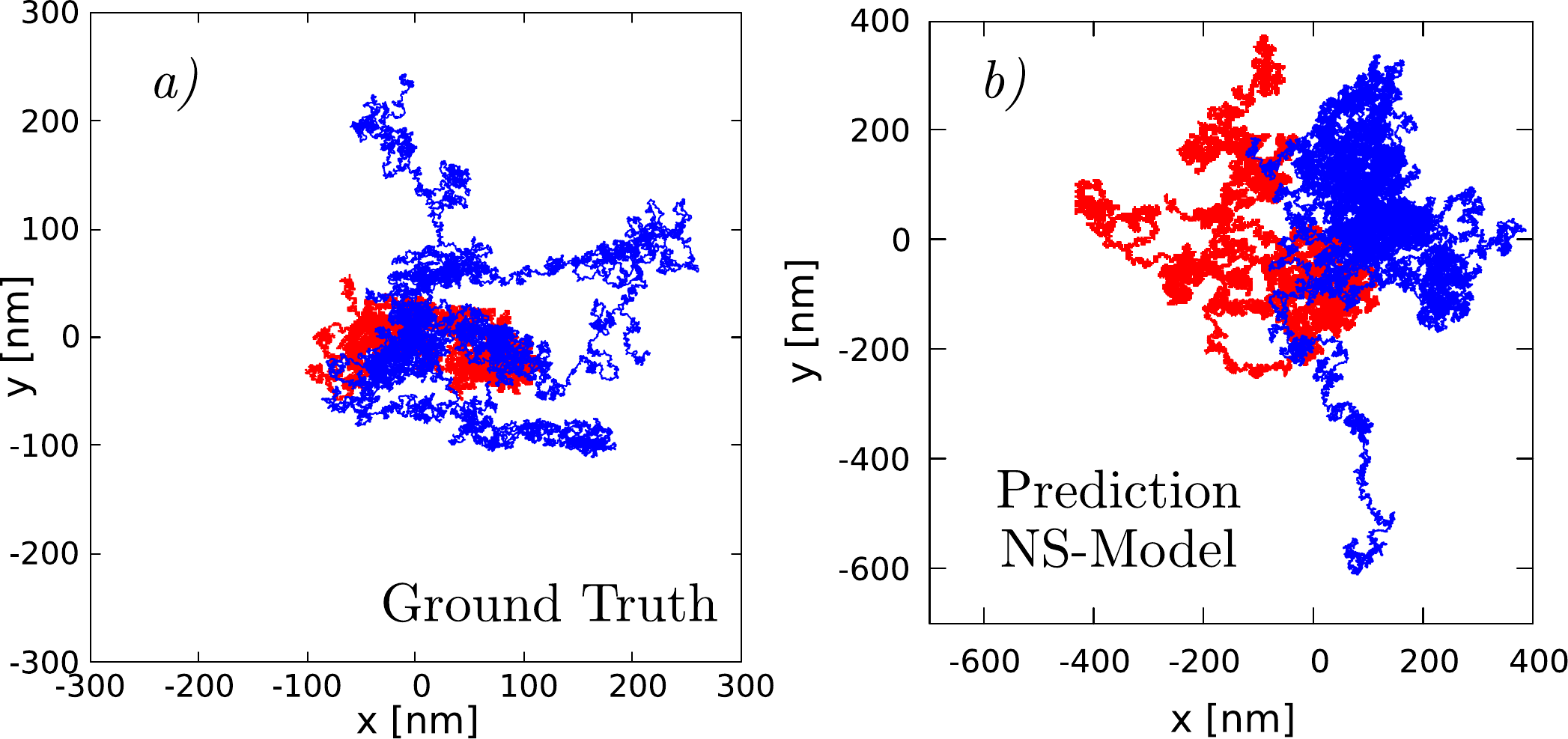}
    \caption{Comparison between the ground truth (MD simulations) and predictions of our neural stochastic (NS) model , for the 7 PIEZO and 10 TREK multi-protein system. \textit{a)} Trajectories of the 7 PIEZO channels (red lines) and 10 TREK channels (blue lines) obtained from MD simulations during a total time of 9,000 time frames. \textit{b)} Predicted trajectories using the NS model of the PIEZO channels and TREK during a total time with 40,000 time frames.      }
    \label{fig:7P10T}
\end{figure}

As we mentioned we also implemented our new NS model for describing and predicting the diffusive dynamics of a multi-protein with two different species of channels. For the 7 PIEZO and 10 TREK system the MD trajectories and those  predicted by our model are shown in Fig.~\ref{fig:7P10T}, displaying the ground truth's tracks in Fig.~\ref{fig:7P10T}~\textit{a)} for MD simulations with 9,000 time frames. And in  Fig.~\ref{fig:7P10T}~\textit{b)}  we display the predicted trajectories using the NS model for 40,000 time frames. In both cases the 7 PIEZO trajectories are shown in red lines and the 10 TREK tracks are shown in blue lines. As we can see in Fig.~\ref{fig:7P10T} Brownian-like dynamics is displayed in both cases, still there are differences in between different protein dynamics.  For instance  PIEZO channels tend to make shorter displacements when compared to TREK trajectories, this effect is more evident in the MD simulations.

\begin{figure}
    \centering
    \includegraphics[width=0.9\linewidth]{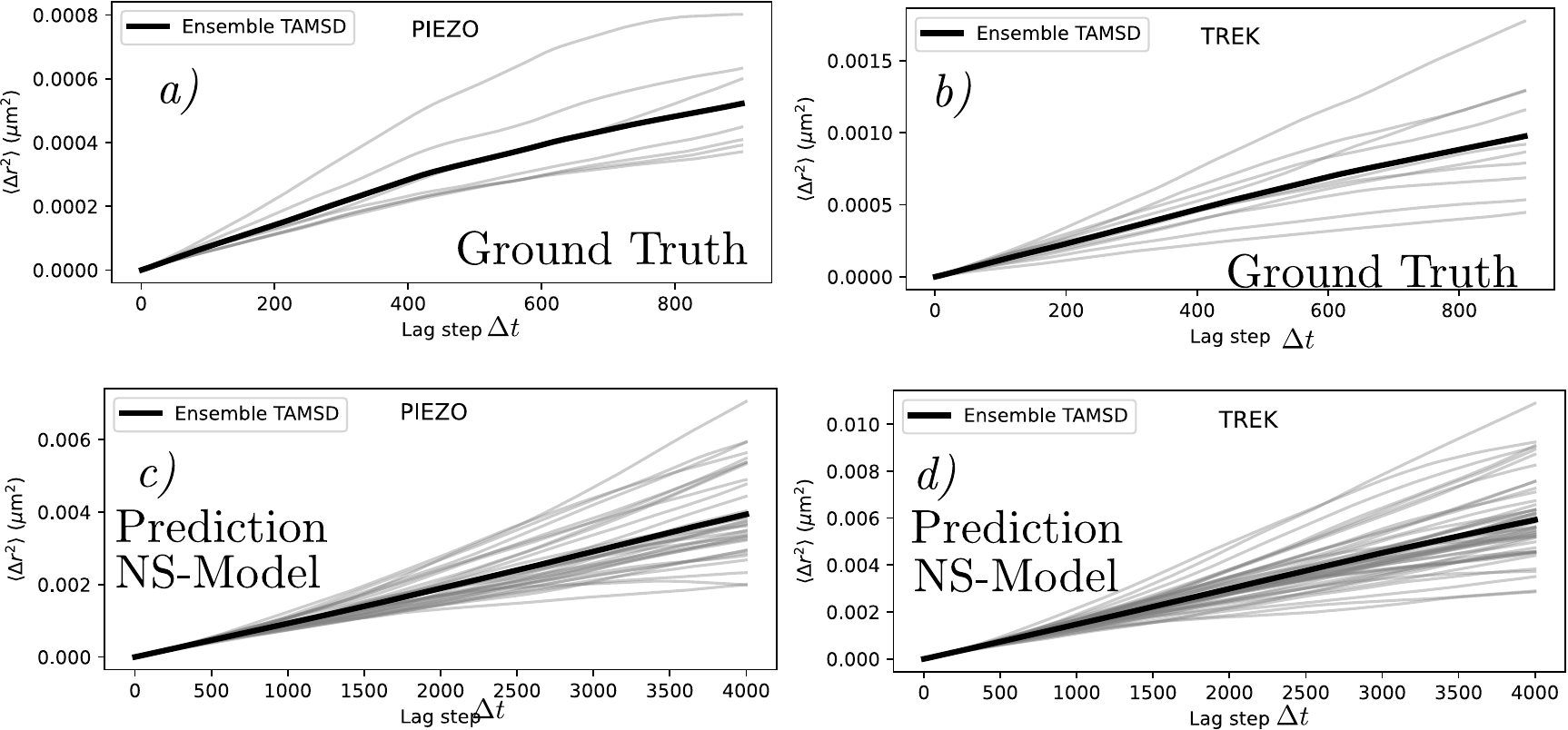}
    \caption{TAMSD:  comparison between the ground truth (MD simulations) and predictions of our neural stochastic (NS) model , for the 7 PIEZO and 10 TREK multi-protein system. \textit{a-b)} Ground Truth TAMSD (gray lines) and ETAMSD (black line) for the 7 PIEZO channels and the 10 TREK proteins respectively. \textit{c-d)} TAMSD and ETAMSD for the NS-Model for the 5 rollouts of the 7PIEZO and 10 TREK system, with 35 TAMSD curves for the PIEZO channels see \textit{c)} and 50 TAMSD curves for TREK channels see \textit{d)}.       }
    \label{fig:TAMSD7P10T}
\end{figure}

Now as we observe in Fig.~\ref{fig:TAMSD7P10T} the TAMSD Eq.~\ref{eq:TAMSD} analysis shows linear dependence of the TAMSD and ETAMSD with respect to the lag time.  Having that  in all cases the ETAMSD renders anomalous exponents close to one, indicating Brownian dynamics in the MD simulations and NS predictions. Particularly  for the PIEZO channels in the Ground Truth we have $\bar{\alpha}=0.943 \pm 0.016$ (SEM). In the case of the TREK channels in the Ground Truth the mean anomalous exponent is $\bar{\alpha}=0.995 \pm 0.019$ (SEM). For the NS predictions the PIEZO channels gave an anomalous exponent of $\bar{\alpha}=1.015 \pm 0.013$ (SEM) and $\bar{\alpha}=1.011 \pm 0.011$ (SEM) for the TREK trajectories. As we note the difference in the values of the anomalous exponent between the ground truth and the NS predictions is covered by the SEM in all cases, so we can say that the NS model predicts accurately the diffusion regime (normal diffusion) even in a different specie multi-protein system.

\begin{figure}
    \centering
    \includegraphics[width=0.9\linewidth]{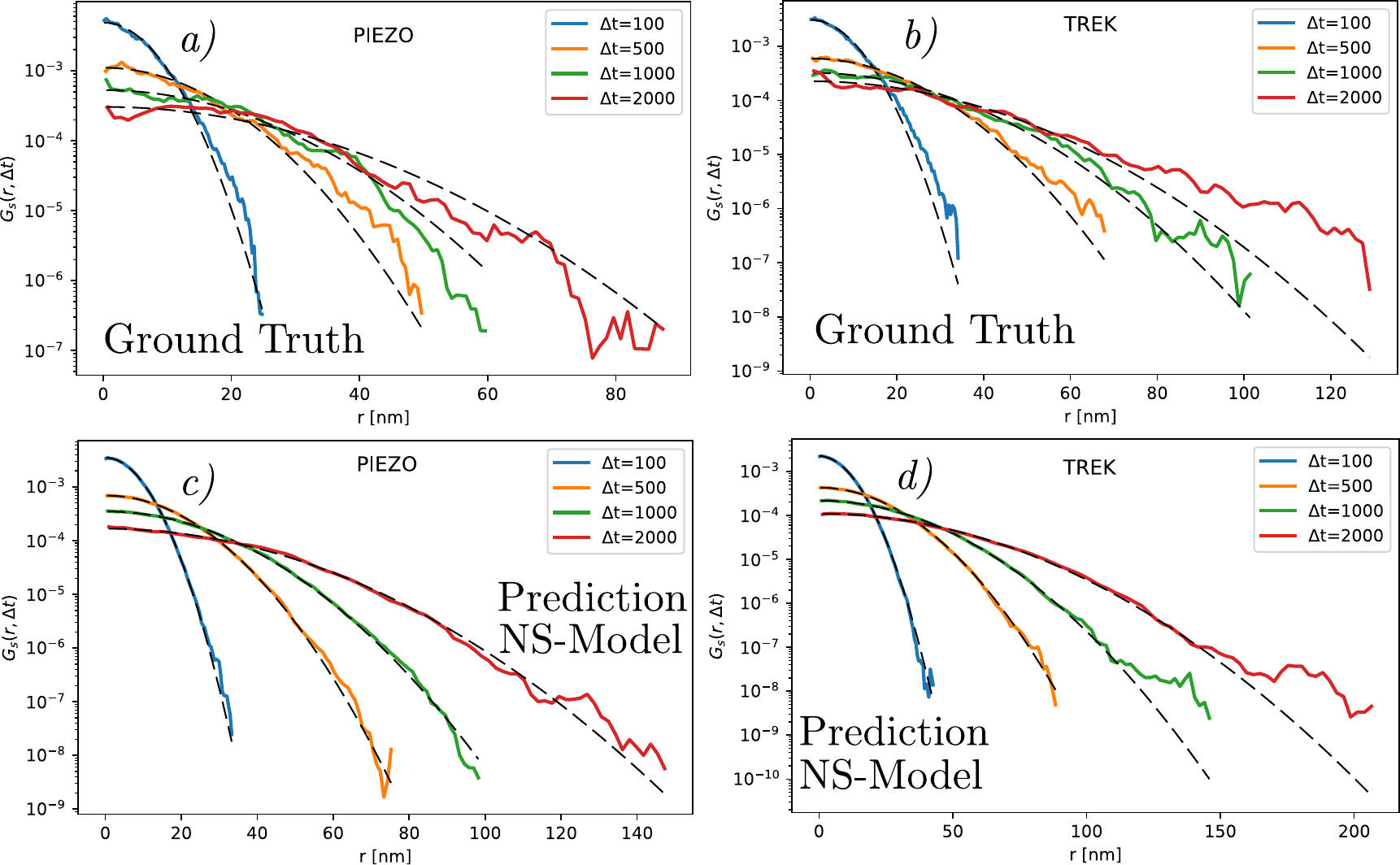}
    \caption{Van Hove function:  comparison between the ground truth (MD simulations) and predictions of our neural stochastic (NS) model , for the 7 PIEZO and 10 TREK multi-protein system. \textit{a-b)} Ground Truth Van Hove function  for the 7 PIEZO channels and the 10 TREK proteins respectively. \textit{c-d)} Van Hove function for the NS-Model for the 5 rollouts of the 7PIEZO and 10 TREK system, for the PIEZO channels see \textit{c)} and  for TREK channels see \textit{d)}.       }
    \label{fig:VH7P10T}
\end{figure}

\begin{figure}
    \centering
    \includegraphics[width=0.9\linewidth]{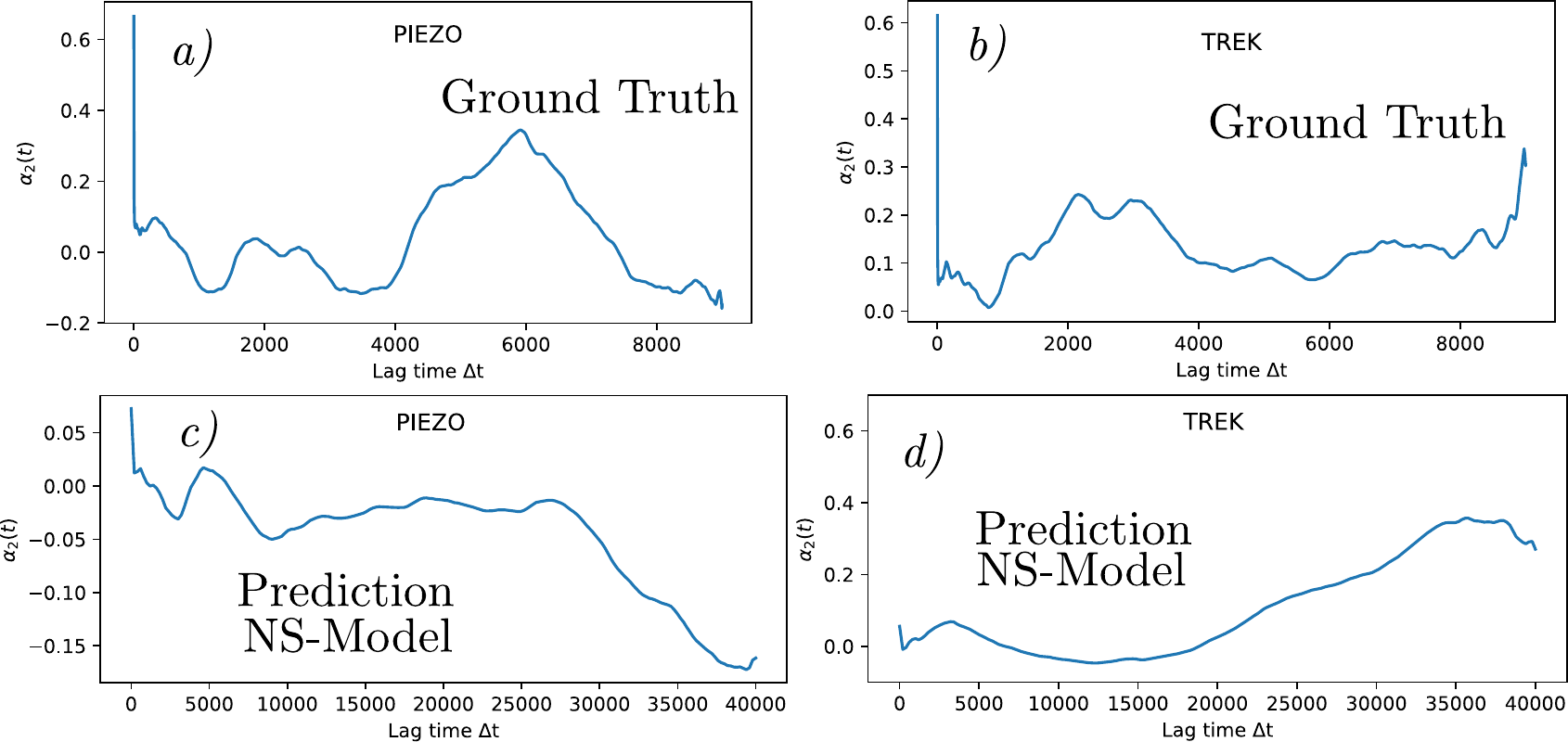}
    \caption{Non Gaussian parameter:  comparison between the ground truth (MD simulations) and predictions of our neural stochastic (NS) model , for the 7 PIEZO and 10 TREK multi-protein system. \textit{a-b)} Ground Truth TAMSD  for the 7 PIEZO channels and the 10 TREK proteins respectively. \textit{c-d)} Non Gaussian parameter  for the NS-Model employing  5 rollouts for the 7PIEZO and 10 TREK system, for the PIEZO channels see \textit{c)} and  for TREK channels see \textit{d)}.       }
    \label{fig:NGP7P10T}
\end{figure}

The difference in the spreading of the displacements for PIEZO and TREK as mentioned is present in Fig.~\ref{fig:7P10T}  was also replicated by the NS model and this can be corroborated by the respective Van Hove function Eq.~\ref{eq:vanhove} as we show in Fig.\ref{fig:VH7P10T}.
For the MD simulations for lag times of $\Delta t =\lbrace 100,500\rbrace$ the Van Hove function exhibits large tails when compared to the Gaussian distribution Eq.~\eqref{eq:gaussfit} and for $\Delta t =\lbrace 1000, 2000\rbrace$ the distribution of displacements  displacements exhibits suppressed tails with respect to the Gaussian distribution. For TREK tracks in the Ground Truth for $\Delta t = \lbrace 100,500,1000,2000\rbrace$ the distribution of displacements exhibits large tails. For the NS prediction and for the lag times analyzed in the case of PIEZO proteins the Van Hove functions resembles a Gaussian distribution and the for the TREK channels for lag times $\Delta t =\lbrace 100,500\rbrace $ the distribution of displacements is Gaussian, whereas for lag times $\Delta t =\lbrace 1000,2000 \rbrace$ the Van Hove function displays large tails. This behavior is corroborated by the evolution of the non Gaussian parameter Eq.~\eqref{eq:NGP} as we show in Fig.~\ref{fig:NGP7P10T}.

As we show in Fig.~\ref{fig:NGP7P10T}~\textit{a)} the fluctuating behavior of the non Gaussian parameter in PIEZO channels for the MD simulations is preserved for lag times $\Delta t > 2000$, but for $\Delta t > 8000$ remains negative, implying the presence of suppressed tails. In the case of TREK trajectories in the Ground Truth for lag times $\Delta t > 2000$ the non Gaussian parameter is remains positive indicating large tails in the distribution of displacements. This trend is hold by the non Gaussian parameter in the NS predictions.

Overall, the Van Hove functions and the non Gaussian parameter show that the NS model reproduces the contrasting displacement statistics of PIEZO and TREK channels. In particular, it captures the tendency toward suppressed tails in PIEZO displacement distributions and enhanced tails in TREK distributions, preserving the distinct dynamical behavior of both protein populations over extended prediction times.

\section{Discussion}

In this work, we introduced a neural stochastic model for forecasting the long-time diffusion dynamics of membrane proteins from coarse-grained molecular dynamics simulations. A central challenge in this task is that reproducing short-time displacement statistics does not necessarily guarantee accurate long-time dynamics, particularly when predictions are generated autoregressively without further input from the reference trajectories. Our results show that combining a multirate recurrent architecture with a stochastic latent process provides a way to retain temporal correlations over extended rollouts. In particular, the model reproduces the nearly Brownian diffusion regime of the seven-PIEZO system, substantially improving upon the previously developed VAR-SDE architecture~\cite{thai2026}, which exhibits a pronounced departure from the reference dynamics at intermediate and long times.

An important aspect of the proposed model is its ability to describe systems containing multiple interacting protein populations. In the mixed PIEZO-TREK system, the model reproduces the distinct mobilities of the two protein species, despite their simultaneous diffusion within the same membrane environment. The differences observed in their displacement distributions are also reflected in the Van Hove functions and non Gaussian parameters. This suggests that the model captures aspects of the collective stochastic dynamics that cannot be described solely by matching the mean squared displacement. The use of species-specific parameters, together with a cross-attention decoder that allows information to be exchanged between proteins, provides a flexible framework for representing heterogeneous multi-protein systems.

Beyond reproducing the transport statistics of the reference simulations, the ability to generate trajectories substantially longer than those available from MD simulations offers a potential computational advantage. The model generates 40,000 frame trajectories, extending beyond the duration of the reference trajectories while preserving several key statistical properties of the underlying dynamics. Such long rollouts could be useful for investigating transport observables that require extensive trajectory sampling, including displacement distributions, temporal correlations, and long-time diffusion behavior. Nevertheless, the generated trajectories should be regarded as statistical surrogates of the simulated dynamics rather than exact reconstructions of individual molecular trajectories or replacements for the underlying molecular dynamics.

Several limitations should also be considered. First, the model is trained and evaluated on a limited set of coarse-grained membrane systems, and its ability to generalize to different protein compositions, membrane properties, and simulation conditions remains to be established. Second, although the architecture incorporates periodic boundary conditions, pairwise geometric information, and stochastic latent dynamics, it does not explicitly enforce the underlying molecular force field or membrane energetics. Its predictions therefore depend on the statistical information captured during training. Future work could investigate the inclusion of additional physical constraints and the evaluation of the model across a broader range of membrane systems. These developments would help establish the extent to which neural stochastic models can provide reliable and computationally efficient tools for studying long-time membrane-protein dynamics.

\backmatter

\bmhead{Acknowledgments}
This work was supported by NIH grant R01HL179692 to Y.L.L. The authors wish to acknowledge the computational resources provided by Stampede3 at the Texas Advanced Computing Center through allocation MCB160119 and BIO240033 from the Advanced Cyberinfrastructure Coordination Ecosystem: Services \& Support (ACCESS) program, which is supported by National Science Foundation grants \#2138259, \#2138286, \#2138307, \#2137603, and \#2138296, and the computational resources available through the National Research Platform (NRP) at the University of California, San Diego. NRP has been developed, and is supported in part, by funding
from National Science Foundation, from awards 1730158, 1540112, 1541349, 1826967,
2112167, 2100237, and 2120019, as well as additional funding from community partners.
\bibliography{sn-bibliography}

\end{document}